\pdfoutput=1

\documentclass[11pt]{article}

\usepackage[]{acl}

\usepackage{times}
\usepackage{latexsym}
\usepackage{subcaption}
\usepackage[T1]{fontenc}

\usepackage[utf8]{inputenc}

\usepackage{microtype}
\usepackage{inconsolata}
\usepackage{booktabs}
\usepackage{multirow}
\usepackage{graphicx}
\usepackage{longtable}
\usepackage{ragged2e}
\usepackage{comment}
\usepackage[most]{tcolorbox}
\usepackage{amsmath}
\usepackage{amssymb}
\usepackage{amsfonts}
\usepackage{bbm}
\usepackage{float}
\usepackage{graphicx}
\usepackage{cuted}
\usepackage{adjustbox}
\usepackage{color}
\usepackage{multicol}
\usepackage{lipsum}
\title{From Evidence to Belief: \\A Bayesian Epistemology Approach to Language Models}

\author{Minsu Kim, Sangryul Kim, James Thorne\\
KAIST AI\\
\texttt{\{minsu\_kim, sangryul, thorne\}@kaist.ac.kr}}

\begin{document}
\maketitle
\begin{abstract}
This paper investigates the knowledge of language models from the perspective of Bayesian epistemology. We explore how language models adjust their confidence and responses when presented with evidence with varying levels of informativeness and reliability. To study these properties, we create a dataset with various types of evidence and analyze language models' responses and confidence using verbalized confidence, token probability, and sampling. We observed that language models do not consistently follow Bayesian epistemology: language models follow the Bayesian confirmation assumption well with true evidence but fail to adhere to other Bayesian assumptions when encountering different evidence types. Also, we demonstrated that language models can exhibit high confidence when given strong evidence, but this does not always guarantee high accuracy. Our analysis also reveals that language models are biased toward golden evidence and show varying performance depending on the degree of irrelevance, helping explain why they deviate from Bayesian assumptions.
\noindent{The Code and data are available at:} \url{https://github.com/MS0117/BayesianEpistemology}
\end{abstract}

\section{Introduction} \label{sec:intro}

Large Language models (LLMs) have advanced to the point where they can naturally 
respond to various practical tasks such as question-answering, code generation and conversation \citep{openai2023gpt4,geminiteam2024gemini}. However, limitations like hallucination and trustworthiness still exist, and research efforts continue to address these issues \citep{Huang2023ASO,sun2024trustllm,Xiao2021OnHA,Zhang2023EnhancingUH}. In this paper, we take a different approach by examining large language models from a philosophical perspective: we investigate whether language models can be said to possess knowledge. 
In epistemology, knowledge is traditionally analyzed using three conditions—truth, justification, and belief—often associated with the justified true belief (JTB) framework \citep{Audi1997-AUDEAC-3}. 
\begin{figure*}[hbt!]

  \centering

  \includegraphics[width=\textwidth]{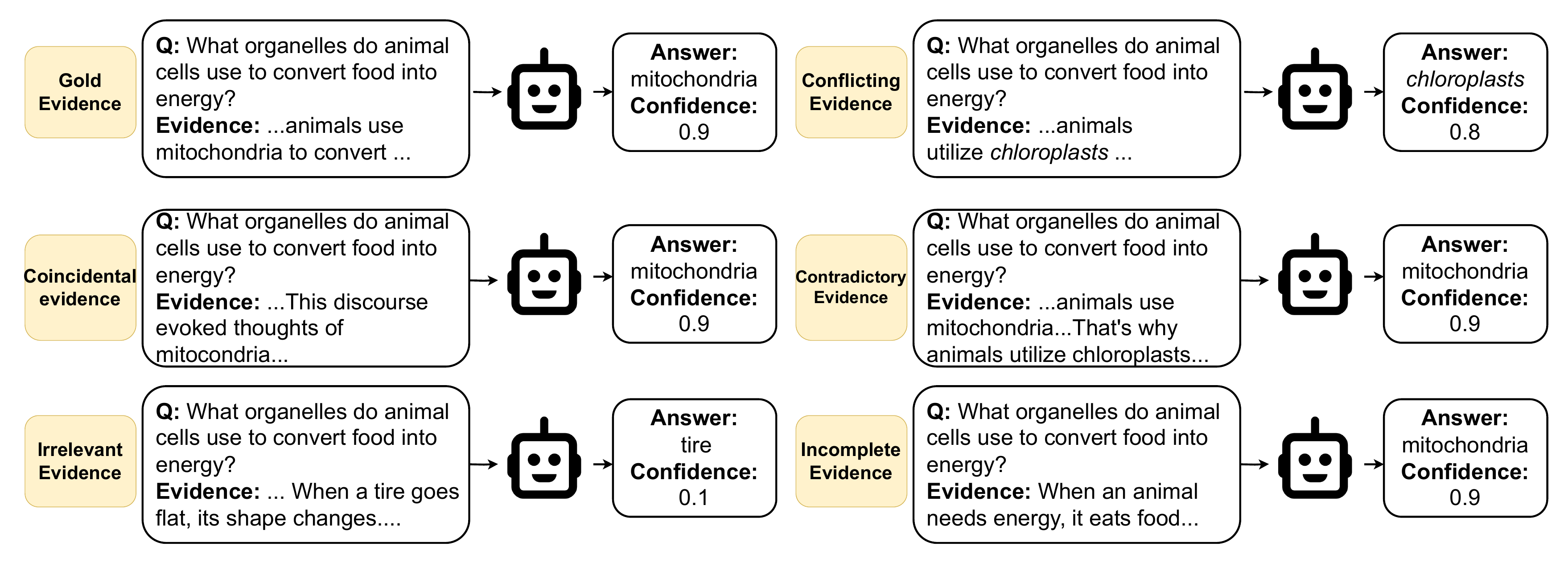}

 \captionsetup{}

  \caption{The overall experimental method and simple examples of the evidence dataset for Confirmation Task. As golden evidence that aligns with the question is given to language models, it shows high confidence and accuracy. However, if language models encounter irrelevant evidence, it responds with low confidence. Further results and analysis are reported in Section \ref{subsec:confirm}. }

  \label{fig:intro}

\end{figure*}
Prior NLP research has focused on two aspects: factual correctness (i.e. the \textit{truth} condition) -- assessing whether the response of a model is correct \citep{hendrycks2021measuring,srivastava2023imitation} -- and \textit{justification}, which encompasses explanation generation  \citep{wei2023chainofthought,camburu2018esnli} and evidence finding \citep{thorne-etal-2018-fever}.

In this work, we investigate whether the model believes its own responses (the \textit{belief} condition): specifically, the relationship between belief and the language model's justification, expressed as evidence. Since belief is a challenging concept to define, this paper focuses on belief from the perspective of Bayesian epistemology, which interprets belief as a quantitative and functional variable.
According to Bayesian epistemology, the degree of belief can be interpreted and measured as probability, called \textit{probability norm}. In particular, regarding the confirmation of belief, we should adjust the confidence of belief based on evidence.  
Specifically, when \( H \) represents the hypothesis (or belief),  \( E \), the evidence for the belief, and \( \theta \) representing the background information or prior knowledge, we can define 3 primitive assumptions:\footnote{According to \citet{Hajek2003-HJEWCP,Vassend2023-VASWHE}, we treat conditional probability as a primitive concept representing the likelihood of an event
occurring under certain conditions, rather than relying on the standard ratio formula \( P(H \mid E) = \frac{P(H \cap E)}{P(E)}\). This approach allows us to apply the Bayesian assumption more intuitively, even in cases where \( P(E)=0 \) or with contradictory evidence. For instance, when observing an unlikely event, such as a shark in a freshwater lake, we still make judgments based on the observation despite it contradicting common knowledge. A detailed discussion can be found in \citet{Hajek2003-HJEWCP}.}

\paragraph{Confirmation Assumption:} \( E \) confirms \( H \) if and only if \( P(H \mid E, \theta) > P(H \mid \theta) \) 
\paragraph{Disconfirmation Assumption:} \( E \) disconfirms \( H \) if and only if \( P(H \mid \theta) > P(H \mid E, \theta) \).
\paragraph{Irrelevance Assumption:} \( E \) is irrelevant to \( H \) if and only if \( P(H \mid \theta) = P(H \mid E, \theta) \).

Also, if we define belief in terms of probability, the strength of the evidence should also be reflected in the confidence. 
That is, 
\begin{itemize}
     \item \textbf{Evidence Power Assumption: }\( E' \) confirms \( H \) more strongly  than \( E'' \)  if and only if \( P(H \mid E', \theta)\) > \( P(H \mid E'', \theta)\)
\end{itemize} 
\citep{Horwich1982-HORPAE-5,Howson2000-HOWHPI,Talbott2006-TALBE,Hajek2010-HJEBE}.
The degrees of belief should not only be a probability. The probabilities assigned to these beliefs must align with the \textit{calibration norm}, meaning they should correspond to the actual likelihood of the event occurring, that is, the actual frequency \citep{Williamson2010-WILIDO-10}. 

The goal of this paper is to explore whether different types of evidence are reflected in language models' confidence and responses.  The evidence here is not merely perturbations altering the correctness of information, i.e., informativeness, but our dataset also includes variations and modifications for reliability factors such as coincidence, timeliness, source of credibility, etc.

Our paper shows that language models can exhibit high confidence and accuracy when encountering true evidence but respond inconsistently with conflicting evidence and reduce confidence and accuracy with irrelevant evidence, contrary to Bayesian assumptions. We also found that LLMs are biased toward golden evidence (typical of annotated information that forms part of datasets) and perform differently across the gradient of irrelevance, supporting an understanding of why LLMs deviate from Bayesian epistemology.

\section{Related Works}

\paragraph{Calibration of LLMs}
Calibration of language models is a key metric for ensuring faithful responses, with log probabilities often representing model confidence \citep{kadavath2022language,lee-etal-2023-large,guo2017calibration}. As models have scaled, research has expanded to verbalized confidence, where models generate their own confidence \citep{lin2022teaching,mielke-etal-2022-reducing,tian2023just}. While confidence can improve model performance \citep{zhao2023slichf,tian2023finetuning}, some studies focus on interpreting this confidence as measuring uncertainty in semantic space \citep{kuhn2023semantic} and exploring confidence through prompt and sampling methods \citep{xiong2024llms}.

\citet{zhou-etal-2023-navigating} explored how epistemic markers affect calibration. However, unlike their focus on linguistic markers, our work examines how changes in epistemic evidence, containing information on both content and reliability, influence confidence and calibration.
\citet{yu2024informationassociationlanguagemodel} report that logical probability and language model probability can differ. This supports our findings that language models may not incorporate evidence into their responses and confidence.

\paragraph{Adversarial Context}
With in-context learning, studies have examined how few-shot demonstrations and explanations affect responses \citep{brown2020language,wei2022finetuned}. \citet{wang2023understanding} showed even inaccurate demonstrations could be used in Chain-of-Thought (COT) prompting, while \citet{chia2023contrastive} improved question accuracy with contrastive demonstrations. \citet{chen2023demonstrations} studied how the number of demonstrations impacts accuracy. \citet{feng2023generictemporalreasoningdifferential} measures language models' responses and probability shifts in temporal relations based on subtle contextual changes. While these works focus on accuracy, we explore how direct question evidence influences not only accuracy but also confidence and calibration.

\citet{turpin2023language,lanham2023measuring} tested perturbations in COT inputs and their impact on answers, similar to our approach. While they focused on modifying explanations based on informativeness (e.g., incorrectness or relevance), our paper investigates whether LLMs reflect diverse evidence in their confidence and calibration, specifically exploring the effects of coincidental evidence and varying source credibility on model confidence.

\section{Methods}
We first generate various types of evidence by few-shot prompting large language models (LLMs) with questions and annotated support from SciQ \citep{welbl-etal-2017-crowdsourcing}, TriviaQA \citep{joshi-etal-2017-triviaqa}, GSM8K \citep{cobbe2021training}. Refer to Appendix \ref{subsec:promptevidence} for details. We then used this evidence to evaluate how the models' confidence and responses change based on the type of evidence provided as shown in Figure \ref{fig:intro}. Influenced by Bayesian epistemology, we defined a confirmation task to measure whether language models can reflect the confirmation, disconfirmation, or irrelevance assumption introduced in Section \ref{sec:intro}. Also, we created a strength-of-evidence task to assess LLM's ability to represent the various power of evidence. To measure the probability norm for adjusting confidence according to the evidence, we used an average confidence across all samples. In order to measure the response, such as correctness or calibration norm, we used accuracy (ACC) and Expected Calibration Error (ECE). In both the confirmation task and the strength-of-evidence task, we used zero-shot prompting for inference.

\subsection{Experimental Design}
We estimated the confidence of language models using verbalized confidence (Verb. 1S top-1) \citep{tian2023just},  token probability, and sampling \citep{lee-etal-2023-large,xiong2024llms}. Refer to Appendix \ref{subsec:eval} and \ref{subsec:promptinference} for details. Smaller-scale open-source LLMs did not tend to generate responses in the correct format matching the prompt of verbalized confidence. Also, following observations from \citet{tian2023just} that closed-source models are better at generating verbal confidence than open-source models, we used GPT-3.5-turbo-0125 and GPT-4o-2024-05-13 for inference. We used SciQ \citep{welbl-etal-2017-crowdsourcing}, TriviaQA 
\citep{joshi-etal-2017-triviaqa} and GSM8K \citep{cobbe2021training} as the source datasets for our Confirmation task, and used only SciQ dataset for Strength of Evidence task, as a scientific question is suitable for making various degree of reliable evidence (see Appendix \ref{apdx:exepriment} for experimental details and dataset statistics).

\subsection{Confirmation Task} \label{subsec:def_confirm}
The objective of the confirmation task to observe and analyze the changes in the language model's confidence and responses when presented with various types of evidence, compared to scenarios where the language models receive the original evidence, \(E\), or in the absence of evidence, and assess how these changes align with three assumptions: Confirmation, Disconfirmation, and Irrelevance introduced section \ref{sec:intro}. Let the entire dataset be
\begin{equation}
\begin{split}
D = \{ & (Q_i, A_i, E_i) \mid \, Q_i \text{ is a question, } \\
& A_i \text{ is an answer for } Q_i, \\
& \text{and } E_i \text{ is evidence for } Q_i \text{ and } A_i \}.
\end{split}
\end{equation}
and 
\begin{equation}
E_i = (s_{i1}, s_{i2}, \ldots, s_{in}) 
\end{equation}
where \( s_{ij} \) is an evidence  sentence in the collection of sentences \( E_i \) indexed by \(j=\{1,\dots,n\}\). For the experiment, we need to create modified \((Q_i, A_i, E'_i)\) where \( E'_i \) is a perturbation of \( E_i \). The following are the types of \( E'_i \):

\begin{enumerate}
    \item \textbf{Conflicting Evidence} \\
    Conflicting evidence refers to abnormal information that hinders reaching the correct answer, introducing misinformation or conflicting beliefs with the golden evidence \(E_i\).
    Specifically, evidence where \( s_{ij} \) in \( E_i \) are replaced with their conflicting counterpart sentences \(\tilde{s}_{ij} \). Thus, \( E'_i \) is conflicting evidence if and only if  
    \[\small
    E'_i = (\tilde s_{i1}, \tilde s_{i2}, \ldots, \tilde s_{in}), \quad \\
    \forall s_{ij} \in E_i.
    \]   
    \item \textbf{Incomplete Evidence} \\
    Evidence that includes only a subset of sentences from the original evidence collection \( E_i \). Thus, \( E'_i \) is a proper subset of \( E_i \). In our experiments, we discard approximately half the sentences from \(E_i \) (i.e. \( |E'_i| \approx 0.5\times |E_i| \)).

    \item \textbf{Contradictory Evidence} \\
    The original evidence \( E_i \) is concatenated with additional negated sentences from \( E_i \). Thus, \( E'_i \) is contradictory evidence if and only if 
    \[\small
    E'_i = E_i \cup N \quad 
    \text{ where}  
    \quad N \subset \left\{\tilde s_{ij}\mid s_{ij} \in E_i \right\} 
    \]
    such that $|N| = 0.5 \times |E_i|$.  That is, adding 50\% of the conflicting evidence to the original evidence.
    \item \textbf{Irrelevant Evidence} \\
    Irrelevant evidence is  \( E'_i = E_j \) where \( j \neq i \). That is, \( E_i \) is randomly shuffled within the dataset \(D\) so that the evidence \(E_i\) of tuple \((Q_i, A_i, E_i)\) is replaced with evidence \(E_j\) from a different tuple \((Q_j, A_j, E_j)\).

    \item \textbf{Coincidental Evidence} \\
    For the SciQ and TriviaQA dataset, unlike other types of evidence, coincidental evidence does not include incorrect answers but explanations reaching the golden answer by irrational reasoning or epistemic luck. Examples include explanations derived from guessing or vague memories. For GSM8K, coincidental evidence consists of a wrong reasoning process but a correct final answer.
\end{enumerate}
We can see the concise examples of the evidence in Figure \ref{fig:intro}.


\subsection{Strength of Evidence}  \label{subsec:def_strength}

This task differs from the Confirmation task in that it focuses on the strength of evidence. Unlike the modified \(E'\) used in the Confirmation task, the evidence used here includes the correct answer but perturbation of reliability. The goal is to understand how differences in the strength of evidence impact confidence and calibration and assess whether LLMs align with Evidence Power Assumption in section \ref {sec:intro}. For each \((Q_i, A_i)\) pair, two types of perturbation \((Q_i, A_i, E'_i)\) and \((Q_i, A_i, E''_i)\) are created. \(E'_i\) represents more reliable evidence, while \(E''_i\) represents relatively less reliable evidence. The following are the types of evidence:

\begin{enumerate}
    \item \textbf{Source of Credibility} \\
         For each \((Q_i, A_i)\) pair, \(E'_i\) means evidence from a highly reputable and authoritative source, while \(E''_i\) means evidence from an anonymous online post or an individual.

    \item \textbf{Specificity and Detail}
    \\
        This involves varying the detail and specificity of the evidence. Similar to source of credibility, for each \((Q_i, A_i)\), \(E'_i\) is highly detailed evidence, while \(E''_i\) is evidence with general mentions related to the question.

    \item \textbf{Timeliness} \\
 This involves modifying the evidence based on its recency. For each \((Q_i, A_i)\), \(E'_i\) consists of recent findings and experiments, while \(E''_i\) consists of relatively older findings and experiments.

    \item \textbf{Experimental Evidence}
  \\
         For each \((Q_i, A_i)\), \(E'_i\) includes evidence derived from precise and controlled experiments, while \(E''_i\) includes evidence where the answer is observed by a witness without experiments.

\end{enumerate}

\begin{table*}[]
\resizebox{\textwidth}{!}{%
\begin{tabular}{c|c|cccccccc}
\hline
                               & \textbf{Dataset}     &\textbf{Metric}   & \textbf{No\_EVI} & \textbf{EVI}   & \textbf{Coincidence} & \textbf{Irrelevant} & \textbf{Conflict} & \textbf{Incomplete} & \textbf{Contradiction} \\ \hline \hline
\multirow{9}{*}{\textbf{GPT-3.5-turbo}} & \multirow{3}{*}{SciQ}  & Confidence             & 0.851   & 0.943 & 0.835       & 0.714      & 0.827    & 0.928      & 0.945          \\
                               &                         & Accuracy \(\uparrow\)   & 0.67    & 0.841 & 0.854       & 0.53       & 0.572    & 0.77       & 0.847         \\
                               &                         & ECE \(\downarrow\)         & 0.18    & 0.111 & 0.071       & 0.262      & 0.304    & 0.161      & 0.108         \\ \cline{2-10} 
                               & \multirow{3}{*}{Trivia} & Confidence            & 0.827   & 0.922 & 0.818       & 0.69       & 0.797    & 0.897      & 0.925         \\
                               &                         & Accuracy \(\uparrow\)  & 0.846   & 0.879 & 0.971       & 0.698      & 0.702    & 0.86       & 0.869         \\
                               &                         & ECE \(\downarrow\)          & 0.035   & 0.058 & 0.153       & 0.125      & 0.211    & 0.06       & 0.076         \\ \cline{2-10} 
                               & \multirow{3}{*}{GSM8K}  & Confidence           & 0.74    & 0.998 & 0.988       & 0.765      & 0.931    & 0.96       & 0.949         \\
                               &                         & Accuracy \(\uparrow\)  & 0.078   & 0.951 & 0.843       & 0.066      & 0.023    & 0.666      & 0.777         \\
                               &                         & ECE \(\downarrow\)         & 0.662   & 0.048 & 0.148       & 0.699      & 0.911    & 0.307      & 0.197         \\ \hline
\multirow{9}{*}{\textbf{GPT-4o}}        & \multirow{3}{*}{SciQ}   & Confidence            & 0.925   & 0.986 & 0.902       & 0.861      & 0.875    & 0.948      & 0.977         \\
                               &                         & Accuracy \(\uparrow\)   & 0.73    & 0.915  & 0.88        & 0.7        & 0.675    & 0.82       & 0.905          \\
                               &                         & ECE \(\downarrow\)        & 0.195   & 0.073 & 0.04       & 0.171      & 0.2      & 0.128      & 0.072         \\ \cline{2-10} 
                               & \multirow{3}{*}{Trivia} & Confidence            & 0.915   & 0.933 & 0.895       & 0.878      & 0.866    & 0.909      & 0.926         \\
                               &                         & Accuracy \(\uparrow\)   & 0.94    & 0.96  & 0.99        & 0.935      & 0.86     & 0.945      & 0.955         \\
                               &                         & ECE \(\downarrow\)       & 0.037   & 0.027 & 0.095       & 0.063       & 0.048    & 0.036      & 0.037         \\ \cline{2-10} 
                               & \multirow{3}{*}{GSM8K}  & Confidence           & 0.924   & 0.991 & 0.83        & 0.89       & 0.883    & 0.96       & 0.957         \\
                               &                         & Accuracy \(\uparrow\)  & 0.24    & 0.97  & 0.54        & 0.195      & 0.165    & 0.774      & 0.96         \\
                               &                         & ECE \(\downarrow\)       & 0.684   & 0.033 & 0.406       & 0.705      & 0.718    & 0.186      & 0.013         \\ \hline
\end{tabular}%

}
\caption{The result of confirmation task with verbal confidence methods. We used 200 samples for GPT-4o due to the cost limit. NO\_EVI refers the question with no context which means \(P(H \mid \theta) \), serving as baseline. Others are the case of \(P(H \mid E,\theta) \) where evidence appears in the context. EVI refers to the context in which the golden evidence from the dataset is given, while the other evidence types are those mentioned in section \ref{subsec:def_confirm}.}
\label{tab:main_table}
\end{table*}
\section{Results and Analysis}
\subsection{LLMs on Confirmation task} \label{subsec:confirm}
The results of the Confirmation task using verbalized confidence are in Table \ref{tab:main_table}, while the token probability and sampling methods are shown in Tables \ref{tab:main_table_log} and \ref{tab:main_table_sampling} in Appendix \ref{apdx:confirmation}.
In Tables \ref{tab:pvalue_verbal}, \ref{tab:pvalue_token} and \ref{tab:pvalue_sample} located in Appendix \ref{apdx:pvalue_confirmation}, we calculated p-values to compare confidence, accuracy, and ECE between providing no evidence (labeled No\_EVI), original (labeled EVI), and perturbed evidence sets to determine if there were significant differences in model performance across these metrics based on the type of evidence. 
Tables \ref{tab:main_table}, \ref{tab:main_table_log}, and \ref{tab:main_table_sampling}, show the changes in various metrics based on the evidence, and Table \ref{tab:pvalue_verbal},  \ref{tab:pvalue_token}, and  \ref{tab:pvalue_sample}  indicate whether those changes are statistically significant.

\paragraph{LLMs follow confirmation assumption}
In Tables \ref{tab:main_table},  \ref{tab:main_table_log} and  \ref{tab:main_table_sampling}, the  NO\_EVI and EVI column show that when \(E\) is golden evidence that helps confirm the answer, we observe  \( P(H \mid E) > P(H ) \) across all models, datasets and methods we used. In Tables \ref{tab:pvalue_verbal}, \ref{tab:pvalue_token}, and \ref{tab:pvalue_sample}, when golden evidence is provided, the p-value for confidence showed at least a marginal increase, with a significant difference particularly observed in verbal, which align well with the \textit{Confirmation Assumption} of Bayesian epistemology.  
Moreover, both accuracy (ACC) and expected calibration error (ECE) showed improved results when given such confirming evidence, which leads to at least marginal difference in p-value, except for ECE in verbal. This indicates that 
language models have strong confidence and handle information well when the evidence contains purely helpful information for deriving the correct answer. This indicates language models satisfy the probability norm and calibration norm in the confirmation case. 

Unlike SciQ and Trivia, the accuracy significantly improves when evidence is provided, and ECE significantly decreases in GSM8K. It shows that language models have parametric knowledge about SciQ and Trivia datasets and struggle with complex reasoning tasks without explanations and reaffirms the importance of explanation in arithmetic tasks \citep{wei2023chainofthought}.

\paragraph{LLMs inconsistently disconfirm conflicting evidence}
Based on Tables \ref{tab:main_table}, \ref{tab:main_table_log}, and \ref{tab:main_table_sampling}, when conflicting evidence that does not lead to the correct answer is provided, confidence tends to decrease. However, the p-value in Tables \ref{tab:pvalue_verbal}, \ref{tab:pvalue_token}, and \ref{tab:pvalue_sample} reveals that conflicting evidence does not have a significant effect on confidence.
With further investigation, only GPT-4o was found to significantly reduce its confidence with a p-value of 0.003.

On the other hand, accuracy decreased significantly across all confidence methods. ECE  significantly increased only in the verbal and token methods. Low confidence indicates that LLMs do not follow the conflicting evidence to generate an answer, but rather that the conflicting evidence creates confusion with existing parametric knowledge, which leads to lower accuracy and higher ECE.

We made the following assumptions regarding this issue: Conflicting evidence can cause the model's confidence to become inconsistent, with overly low or high confidence appearing regardless of accuracy, which may increase ECE (although low confidence is likely to occur more frequently). Alternatively, conflicting evidence can disrupt the model's learned patterns, making its confidence less reflective of actual accuracy, thereby leading to higher ECE.

In conclusion, only GPT-4o with verbal method exhibited behavior aligned with Bayesian disconfirmation assumptions, showing a decrease in both confidence and accuracy when conflicting evidence was presented.

\begin{figure*}[t!]

  \centering

  \includegraphics[width=\textwidth]{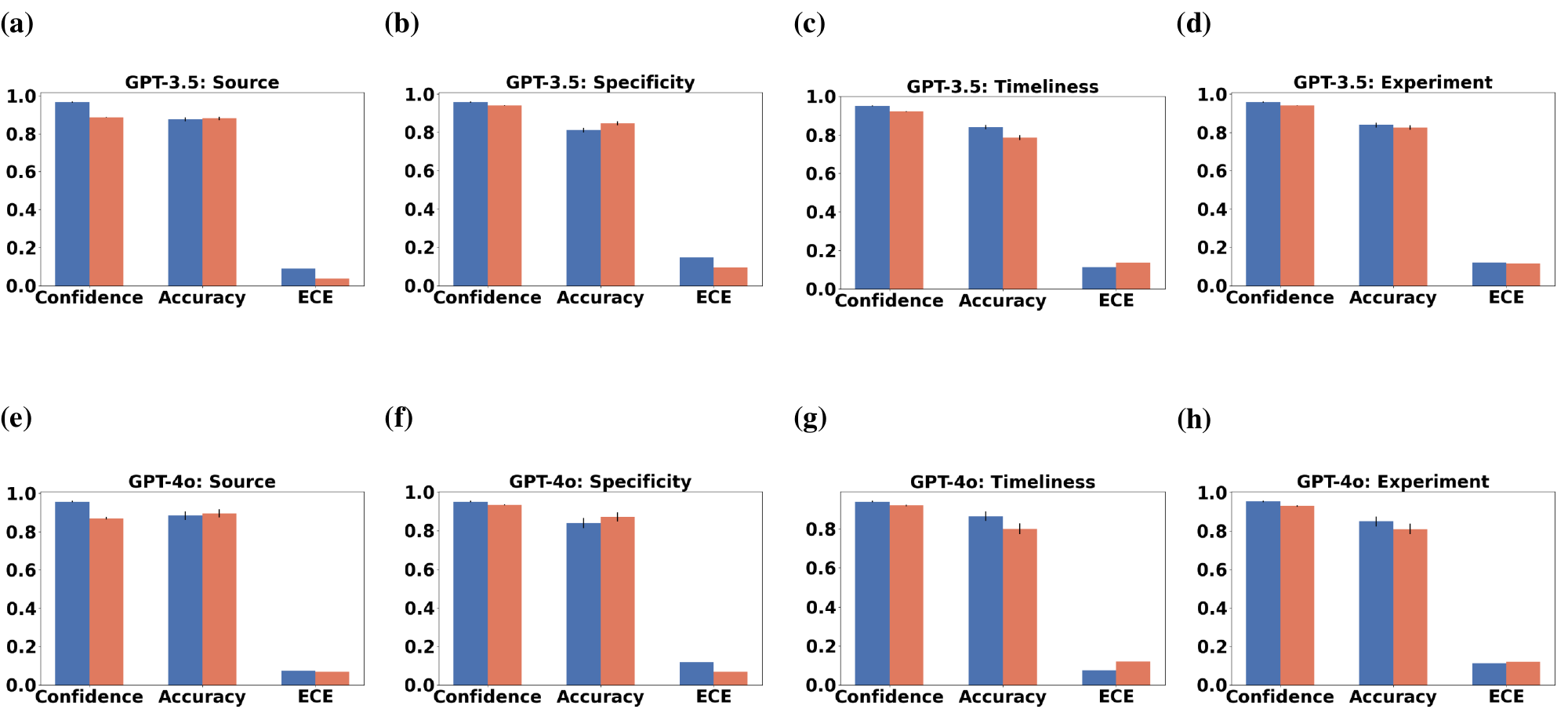}

 \captionsetup{}

  \caption{The results of the Strength of Evidence task on the SciQ dataset with verbal confidence method. The blue bar indicates more credible, specific, recent, and experimental evidence, while the red bar represents less credible, less specific, older, and observational evidence provided to the LLMs. We found that, in all models and datasets, strong evidence leads to high confidence with verbalized confidence. However, it does not always result in improvements in ACC and ECE.
}

  \label{fig:strength}

\end{figure*}
\paragraph{LLMs handle contradictory evidence as golden evidence}
In most models and methods, contradictory evidence, which contains both correct and conflicting evidence in the context, shows increased confidence and accuracy compared to the no-evidence baseline, with p-values close to 0.05 for both cases. Additionally, ECE also decreases with p-value around 0.1. It means that despite the presence of conflicting information, the model appears highly confident and well-calibrated in almost all scenarios, which is similar with golden evidence case. This suggests that LLMs can effectively filter the given context and generate responses without conflicting with their parametric knowledge. Unlike the case with conflicting evidence, it can be interpreted that the influence of golden evidence offsets the presence of incorrect sentences. Hence, LLMs do not consider contradictory evidence as disconfirming their beliefs.
\paragraph{LLMs cannot handle coincidental evidence well.}
When coincidental evidence was provided in both the token and sampling methods, confidence increased marginally (See Tables \ref{tab:main_table_log},  \ref{tab:main_table_sampling}, \ref{tab:pvalue_token}, and \ref{tab:pvalue_sample}). In terms of verbal confidence in Tables \ref{tab:main_table} and \ref{tab:pvalue_verbal}, it was found that there was no significant difference between reliable and unreliable evidence. These indicate that the language model fails to capture the unreliability of evidence and, as a result, cannot properly reflect this in its confidence. However, based on the increased accuracy and its meaningful p-value results, it was revealed that the language model incorporated this unreliable evidence into its response to arrive at the correct answer. The ECE  also decreased for both the token probability method and the sampling method. The general results for token probability and sampling methods suggest that when confidence is measured using these methods, coincidental evidence can have similar effects as golden evidence. On the other hand, there was no significant difference in ECE when using verbal confidence.

In conclusion, the language model appears to have problems with handling unreliable evidence. This is likely because, during training, the model is mostly exposed to correct data and does not frequently encounter incorrect situations.

\paragraph{Incomplete evidence acts as a positive hint.}
Incomplete evidence, though not as strong as the golden evidence case, led to a marginal increase in confidence across all confidence methods. Accuracy and ECE also showed a tendency to marginally increase and decrease, respectively. In most cases, displaying a pattern very similar to that of the golden evidence case. Incomplete evidence does not contain inaccurate information and is a partial subset of the gold evidence, acting as a hint. Similar to the contradictory evidence case, we observe that the language model is biased towards imperfect golden evidence. Therefore, while not as effective as golden evidence, the language model reflects the information from the evidence well without distraction.

\paragraph{LLMs are confused by irrelevant evidence}

Except for the sampling method, when irrelevant evidence was provided, confidence decreased either statistically significantly or marginally. In terms of accuracy, it was found that accuracy significantly decreased. Furthermore, in the verbal method, ECE is also notably increased.

This indicates that language models are severely distracted by irrelevant text in terms of the content of the evidence as in \citet{shi2023large}. These results showed that LLMs do not align with the irrelevance assumption of Bayesian epistemology.
\subsection{LLMs on Strength of Evidence task}
The results of the Strength of Evidence task using the verbalized confidence method, token probability method, sampling method are reported in Figure \ref{fig:strength}, with Figures \ref{fig:strength_token} and \ref{fig:strength_sampling} located in Appendix \ref{apdx:strength}. In Table \ref{tab:pvalue_strength}, we calculated p-values to compare confidence, accuracy, and ECE between more reliable evidence and less reliable evidence to determine if there were significant differences in model performance across these metrics based on the strength of the evidence. Hence, as in section \ref{subsec:confirm}, we can observe the variations in different metrics depending on the power of evidence in Figures \ref{fig:strength}, \ref{fig:strength_token}, and \ref{fig:strength_sampling}, and check the statistical significance in Table \ref{tab:pvalue_strength}.

\paragraph{Strong evidence can give high confidence in Verbal and Sampling methods, but cannot guarantee accurate response}

In Figures \ref{fig:strength}, \ref{fig:strength_sampling} and Table \ref{tab:pvalue_strength}, the confidence increases when more reliable evidence is provided in verbal and sampling methods. Additionally, the p-values for verbal confidence and sampling confidence between weak and strong evidence are 0.015 and 0.038, respectively. This indicates that the model's response confidence significantly increases when strong evidence is provided. 

In verbal methods, as in (a), (b), (e), (f) in Figure \ref{fig:strength}, when low credible source and low detailed evidence were used, accuracy increased and ECE decreased. This suggests that in some cases, strong evidence may not be as useful as we expected for the language model to infer the correct answer. High confidence combined with low accuracy ultimately leads to overconfidence in incorrect predictions, resulting in high ECE. On the other hand, as in (c), (d),(g), (h) in Figure \ref{fig:strength}, evidence containing the latest information or experiments showed higher confidence and accuracy compared to older information or observation-based evidence. Except for GPT-3.5 with experimental evidence, the ECE of stronger evidence was also lower, indicating that using stronger evidence in the cases of timeliness and experiments results in well-calibrated models. This means that in these cases, the language model utilizes the given evidence effectively and accurately reflects the information in its predictions.

The sampling confidence showed higher accuracy when high-reliability evidence is provided in most cases except for specificity.  We consider this phenomenon another positive aspect of self-consistent decoding \citep{wang2023selfconsistency}. A single response might not fully capture the reliability of evidence, such as credibility, timeliness, etc. However, multiple responses can increase the likelihood of accurately reflecting these aspects.

In the case of specificity, both verbalized confidence and sampling failed to adequately to reflect the concreteness of the evidence in the responses. We interpreted that more detailed information can enhance confidence, but it also suggests that such excessive information may hinder the extraction of correct answers that match the question.

\paragraph{Token probability cannot reflect various degrees of reliability.}
As in Figure \ref{fig:strength_token} in Appendix \ref{apdx:strength}, with token probability, confidence did not increase even when stronger evidence was presented. For example, with token probability, when the specificity of the evidence was altered or when the source's credibility was varied in GPT-4o, it failed to reflect confidence according to the strength of the evidence accurately. However, it accurately reflected reliability changes according to the source's credibility, timeliness, and whether an experiment was conducted in the evidence to its accuracy. Additionally, it showed a decrease in ECE in cases of timeliness and experimental evidence.

Through this experiment, we found that when stronger evidence is provided to the language model, it can significantly increase its verbalized and sampling confidence. However, this does not always lead to improvements in accuracy-related performance. 

\section{Ablation}

\begin{figure*}[t!]

  \centering

  \includegraphics[width=\textwidth]{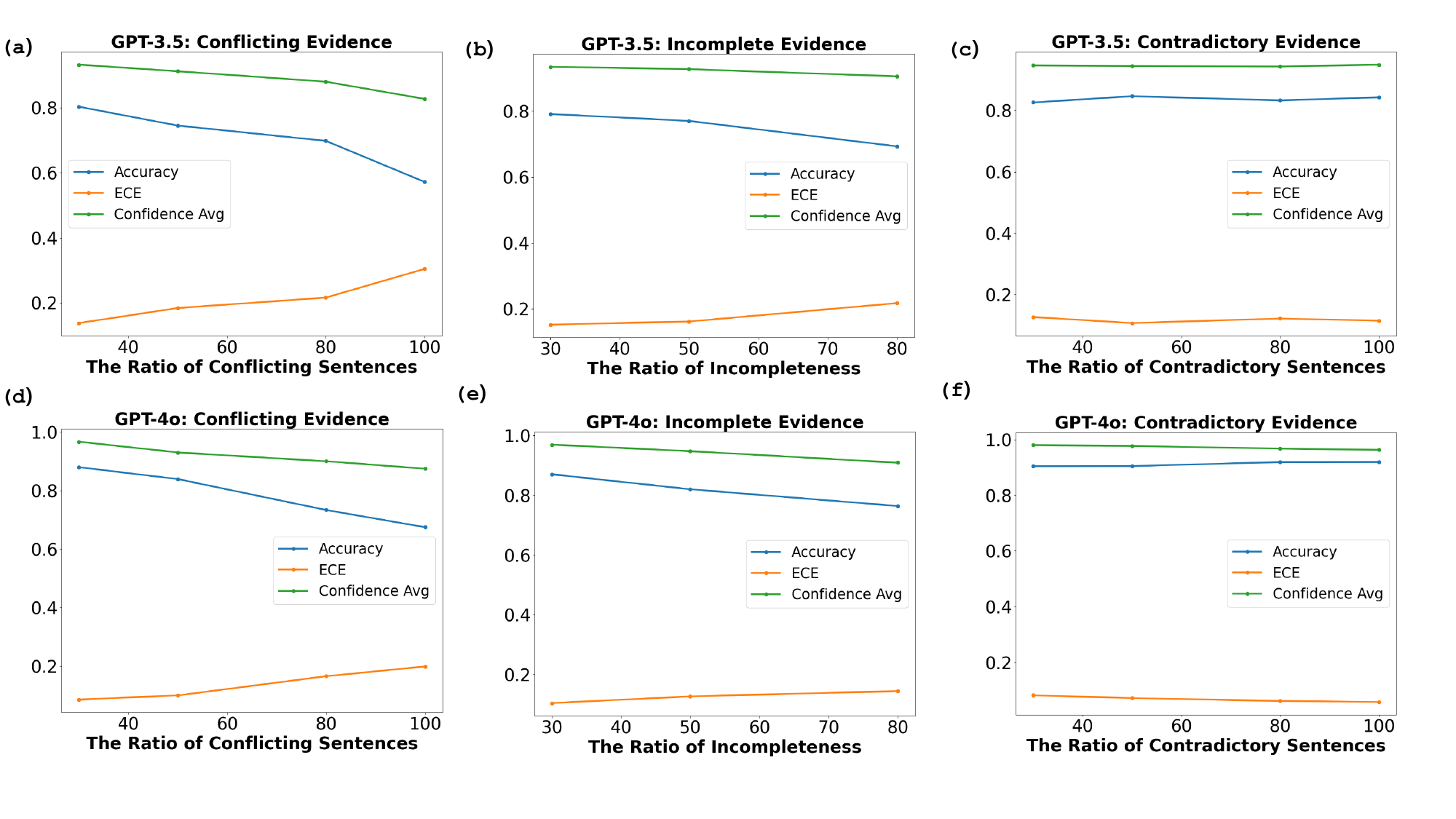}

 \captionsetup{}

  \caption{The results for the degree of variations in evidence for the SciQ dataset with verbalized method. We modified the number of conflicting sentences in conflicting  evidence, sentences in incomplete evidence, and contradictory sentences in contradictory evidence (See Appendix \ref{apdx:golden} for entire results).}

  \label{fig:ablation}

\end{figure*}

\paragraph{LLMs tend to focus more on correct than incorrect information.}
In Section \ref{subsec:confirm}, we interpreted that the language model possesses a certain degree of knowledge about the question in its parameters and tends to be biased towards contexts aligned with this parametric knowledge rather than context hindering it, as seen in golden, contradictory, and incomplete evidence. To justify this, we conducted an experiment adjusting the ratio of golden sentences in conflicting, incomplete, and contradictory evidence. Figure \ref{fig:ablation} (a) and (d) show that as the number of original golden sentences decreases and the conflicting sentence increases, the performance of the language model gradually declines. However, it decreases significantly when there are no golden sentences left. Moreover, Figure \ref{fig:ablation} (b) and (e) demonstrate that as the original golden sentence decreases, performance decreases. On the other hand, Figure \ref{fig:ablation} (c) and (f) indicate that if the golden evidence sentences is sufficiently given,  increasing the number of contradictory sentences does not affect the confidence and performance even if both of the numbers of contradictory sentences and golden evidence sentences are same. This shows that the language model focuses more on the given golden evidence in the context than inaccurate evidence, and this is why it maintains confidence and calibration despite incomplete and contradictory evidence.

\paragraph{Why do LLMs get confused by irrelevant context?}
Two interpretations are possible for the irrelevant case
\begin{enumerate}
    \item The language model does not recognize irrelevant evidence as irrelevant when it is in the same field but differs in content.
    \item The language model considers irrelevant evidence as a kind of noise, which distracts the model and causes confusion.
\end{enumerate}    
To verify (1), instead of extracting irrelevant evidence from the same dataset, we used contexts from different datasets, for SciQ and TriviaQA dataset, we used evidence of GSM8K, and for GSM8K, using TriviaQA. As we can see in Figure \ref{fig:irrelevant} in Appendix \ref{apdx:irrelevant}, even when using a new irrelevant sentence, it did not completely match the completely irrelevant assumption. However, surprisingly, when using evidence from a completely different field, we found that the confidence, accuracy, and ECE metrics approached closer to the baseline no evidence case (P(H)) than when we used evidence where the content was different but the field was the same. This implies that as the irrelevance increases, the LLMs become less distracted by the context. Therefore, we interpreted that there is a possibility that the LLMs satisfy the irrelevant assumption of Bayesian epistemology.

\section{Why Do LLMs Struggle to Follow Bayesian Assumptions?}
We speculate that LLMs may not fully adhere to Bayesian assumptions due to limitations in both their training data and methods. Pretraining datasets, primarily sourced from web crawls with filtering and books \citep{gao2020pile800gbdatasetdiverse}, expose models mostly to well-justified information and correct explanations. This allows them to excel in deriving correct answers but limits their ability to handle coincidental evidence or conflicting beliefs, which are underrepresented in the data.

In addition to data limitations, the training methods used for LLMs diverge significantly from human language acquisition. Humans encounter unreliable and conflicting situations through interaction and experience with the real world. In contrast, LLMs are trained in a largely supervised manner, resulting in a lack of semantic understanding \citep{bender-koller-2020-climbing,bisk-etal-2020-experience,soni-etal-2024-large}. As a result, these limitations may explain why LLMs fail to fully align with Bayesian assumptions when faced with conflicting or unexpected scenarios, as they lack the experiential grounding to properly handle such evidence.

\section{Conclusion}
In this paper, we explored how changes in the informativeness and reliability of evidence affect the confidence and response of language models. 
Specifically, we examined how well language models stick to the probability and calibration norms outlined in Bayesian epistemology. We demonstrated that language models partially align with Bayesian epistemology, following confirmation assumptions but failing to adhere to disconfirmation and irrelevance assumptions. It can be interpreted that language models do not possess a justified belief in the view of Bayesian epistemology. Additionally, we found that LLMs show a bias toward golden evidence and modify their confidence and response relative to the degree of irrelevance, which helps clarify their deviation from Bayesian assumptions. These findings provide philosophical insight into the nature of "belief" in LLMs, highlighting their biases and limitations.

\section{Limitations}
In this paper, we did not theoretically investigate the causes behind the observed phenomena, such as the training algorithm and model architecture, leaving such investigations and their potential implications for future research. One limitation of our dataset is the varying nature of conflicting evidence, some evidence is entirely incompatible, while other types obstruct correct answers. Hence, as we varied the extent of irrelevance in the ablation study, a finer classification of conflicting evidence could benefit future research. Additionally, Bayesian epistemology is not the only theory for defining knowledge. Therefore, our results do not imply a definitive conclusion about whether LLMs possess beliefs or knowledge. 
We did not conduct a human evaluation as a baseline in this study as our focus was on aligning language models with ideal Bayesian epistemology.  Future research could incorporate human evaluation to further assess how models' belief updating and confidence calibration compare to human cognitive processes.
Lastly, further research using more complex and practical datasets, as well as developing new algorithms to address the issues identified in this study, will be valuable in advancing robust AI systems.

\section{Ethics Statement}
In the preparation of this paper, we utilized GPT-4o, for grammatical corrections and coding assistance. This technology served as an auxiliary resource to enhance the clarity and accuracy of our work, without directly influencing the research outcomes or decision-making processes involved. We acknowledge the support provided by OpenAI's GPT-4o in refining the presentation of our findings, ensuring that our use of this tool adheres to ethical guidelines and does not compromise the integrity of our research. 

\section{Acknowledgement}
This work was supported by Institute for Information \& communications Technology Promotion (IITP) grant funded by the Korea government (MSIT) (RS-2019-II190075) Artificial Intelligence Graduate School Program (KAIST) and Artificial intelligence industrial convergence cluster development project funded by the Ministry of Science and ICT (MSIT, Korea) \& Gwangju Metropolitan City.

\bibliography{custom}
\bibliographystyle{acl_natbib}

\newpage
\appendix

\onecolumn
\section*{Appendix}

\bigbreak

\section{Results of Confirmation task}\label{apdx:confirmation}
\begin{table*}[!hbt]
\resizebox{\textwidth}{!}{%
\begin{tabular}{c|c|cccccccc}
\hline
                               & \textbf{Dataset}     &\textbf{Metric}   & \textbf{No\_EVI} & \textbf{EVI}   & \textbf{Coincidence} & \textbf{Irrelevant} & \textbf{Conflict} & \textbf{Incomplete} & \textbf{Contradiction} \\ \hline \hline
\multirow{9}{*}{\textbf{GPT-3.5-turbo}} & \multirow{3}{*}{SciQ}  & Confidence & 0.671   & 0.781 & 0.785       & 0.594      & 0.638    & 0.723     & 0.764          \\
                               &                         & Accuracy \(\uparrow\)  & 0.676    & 0.829 & 0.839       & 0.526      & 0.6   & 0.792       & 0.837         \\
                               &                         & ECE  \(\downarrow\)          & 0.312    & 0.171 & 0.154       & 0.44     & 0.381    & 0.205      & 0.16         \\ \cline{2-10} 
                               & \multirow{3}{*}{Trivia} & Confidence & 0.834   & 0.864 & 0.894       & 0.699      & 0.759    & 0.843      & 0.849         \\
                               &                         & Accuracy \(\uparrow\)  & 0.858   & 0.872 & 0.976       & 0.653      & 0.742    & 0.851      & 0.857         \\
                               &                         & ECE  \(\downarrow\)        & 0.134   & 0.127 & 0.127       & 0.324      & 0.251    & 0.141      & 0.139         \\ \cline{2-10} 
                               & \multirow{3}{*}{GSM8K}  & Confidence & 0.218   & 0.932 & 0.933       & 0.172      & 0.738    & 0.765      & 0.801         \\
                               &                         & Accuracy \(\uparrow\)  & 0.098   & 0.961 & 0.852       & 0.068      & 0.028    & 0.677      & 0.755         \\
                               &                         & ECE  \(\downarrow\)        & 0.777   & 0.046 & 0.148       & 0.725      & 0.939    & 0.299      & 0.222         \\ \hline
\multirow{9}{*}{\textbf{GPT-4o}}        & \multirow{3}{*}{SciQ}   & Confidence & 0.621   & 0.799 & 0.833       & 0.565      & 0.653    & 0.744      & 0.813         \\
                               &                         & Accuracy \(\uparrow\)   & 0.711   & 0.92  & 0.905       & 0.675      & 0.655    & 0.835      & 0.925          \\
                               &                         & ECE   \(\downarrow\)       & 0.276   & 0.082 & 0.1         & 0.314      & 0.334    & 0.165      & 0.078        \\ \cline {2-10} 
                               & \multirow{3}{*}{Trivia} & Confidence & 0.837   & 0.916 & 0.911       & 0.824      & 0.824    & 0.889      & 0.91          \\
                               &                         & Accuracy \(\uparrow\)  & 0.944   & 0.955 & 0.99        & 0.905      & 0.82     & 0.94       & 0.95          \\
                               &                         & ECE  \(\downarrow\)        & 0.06    & 0.047 & 0.01        & 0.088      & 0.173    & 0.064      & 0.05          \\ \cline{2-10} 
                               & \multirow{3}{*}{GSM8K}  & Confidence & 0.354   & 0.865 & 0.54        & 0.299      & 0.372    & 0.755      & 0.842         \\
                               &                         & Accuracy \(\uparrow\)  & 0.249   & 0.97  & 0.505       & 0.227      & 0.191    & 0.83       & 0.955         \\
                               &                         & ECE   \(\downarrow\)       & 0.715   & 0.03  & 0.473       & 0.697      & 0.74     & 0.157      & 0.037         \\ \hline
\end{tabular}%

}
\caption{The result of confirmation task with token probability method. We used 200 samples for GPT-4o due to the cost limit. NO\_EVI refers the question with no context which means \(P(H \mid \theta) \), serving as baseline. Others are the case of \(P(H \mid E,\theta) \) where evidence appears in the context. EVI refers to the context in which the golden evidence from the dataset is given, while the other evidence types are those mentioned in section \ref{subsec:def_confirm}.}
\label{tab:main_table_log}
\end{table*}
\begin{table*}[!hbt]
\resizebox{\textwidth}{!}{%
\begin{tabular}{c|c|cccccccc}
\hline
                               & \textbf{Dataset}     &\textbf{Metric}   & \textbf{No\_EVI} & \textbf{EVI}   & \textbf{Coincidence} & \textbf{Irrelevant} & \textbf{Conflict} & \textbf{Incomplete} & \textbf{Contradiction} \\ \hline \hline
\multirow{9}{*}{\textbf{GPT-3.5-turbo}} & \multirow{3}{*}{SciQ}  & Confidence  & 0.874   & 0.921 & 0.916       & 0.798      & 0.828    & 0.888      & 0.922         \\
                               &                         & Accuracy \(\uparrow\)  & 0.693   & 0.846 & 0.853       & 0.551      & 0.617    & 0.777      & 0.853         \\
                               &                         & ECE  \(\downarrow\)        & 0.18    & 0.076 & 0.077       & 0.248      & 0.211    & 0.111      & 0.074          \\     \cline{2-10} 
                               & \multirow{3}{*}{Trivia} & Confidence & 0.921   & 0.939 & 0.963       & 0.822      & 0.862    & 0.924      & 0.934         \\
                               &                         & Accuracy \(\uparrow\)  & 0.869   & 0.884 & 0.979       & 0.668      & 0.693    & 0.856      & 0.884         \\
                               &                         & ECE  \(\downarrow\)        & 0.057   & 0.059 & 0.034       & 0.154      & 0.17     & 0.072      & 0.076         \\ \cline{2-10} 
                               & \multirow{3}{*}{GSM8K}  & Confidence & 0.422   & 0.986 & 0.977       & 0.377      & 0.838    & 0.86       & 0.848         \\
                               &                         & Accuracy \(\uparrow\)  & 0.12    & 0.967 & 0.849       & 0.059      & 0.028    & 0.716      & 0.756         \\
                               &                         & ECE   \(\downarrow\)       & 0.302   & 0.036 & 0.138       & 0.318      & 0.81     & 0.144      & 0.091         \\  \hline
\multirow{9}{*}{\textbf{GPT-4o}}        & \multirow{3}{*}{SciQ}   & Confidence & 0.872   & 0.968 & 0.959       & 0.852      & 0.871    & 0.923      & 0.965         \\
                               &                         & Accuracy \(\uparrow\)  & 0.694   & 0.934 & 0.924       & 0.708      & 0.698    & 0.84       & 0.933         \\
                               &                         & ECE \(\downarrow\)         & 0.18    & 0.06  & 0.102       & 0.149      & 0.114    & 0.132      & 0.066        \\ \cline {2-10} 
                               & \multirow{3}{*}{Trivia} & Confidence & 0.845   & 0.973 & 0.973       & 0.943      & 0.918    & 0.966      & 0.97          \\
                               &                         & Accuracy \(\uparrow\)  & 0.945   & 0.969 & 0.99        & 0.924      & 0.843    & 0.924      & 0.959         \\
                               &                         & ECE  \(\downarrow\)        & 0.053   & 0.026 & 0.016       & 0.04       & 0.122    & 0.042      & 0.038         \\ \cline{2-10} 
                               & \multirow{3}{*}{GSM8K}  & Confidence & 0.506   & 0.958 & 0.684       & 0.481      & 0.529    & 0.875      & 0.957         \\
                               &                         & Accuracy \(\uparrow\)  & 0.3     & 0.969 & 0.587       & 0.257      & 0.224    & 0.829      & 0.969         \\
                               &                         & ECE  \(\downarrow\)        & 0.206   & 0.065 & 0.156       & 0.224      & 0.305    & 0.103      & 0.051         \\ \hline
\end{tabular}%

}
\caption{The result of confirmation task with sampling method. We used 200 samples for GPT-4o due to the cost limit. NO\_EVI refers the question with no context which means \(P(H \mid \theta) \), serving as baseline. Others are the case of \(P(H \mid E,\theta) \) where evidence appears in the context. EVI refers to the context in which the golden evidence from the dataset is given, while the other evidence types are those mentioned in section \ref{subsec:def_confirm}.}
\label{tab:main_table_sampling}
\end{table*}

\clearpage
\bigbreak

\section{Results of p-value for  Confirmation task}\label{apdx:pvalue_confirmation}
We consider $p \leq 0.05$ as statistically significant, while $0.05 < p \leq 0.1$ is regarded as marginally significant.

\begin{table*}[!hbt]
\resizebox{\textwidth}{!}{%
\begin{tabular}{l|cccccc}
\hline
           & \multicolumn{1}{l}{\textbf{NO\_EVI-EVI}} & \multicolumn{1}{l}{\textbf{NO\_EVI-Coincidence}} & \multicolumn{1}{l}{\textbf{NO\_EVI-IRR}} & \multicolumn{1}{l}{\textbf{NO\_EVI-Conflict}} & \multicolumn{1}{l}{\textbf{NO\_EVI-Incomplete}} & \multicolumn{1}{l}{\textbf{NO\_EVI-Contradiction}} \\ \hline
Confidence & 0.033                            & 0.779                                   & 0.059                           & 0.99                            & 0.084                                  & 0.035                                     \\
ACC        & 0.077                           & 0.056                                   & 0.066                           & 0.002                           & 0.097                                  & 0.071                                     \\
ECE        & 0.114                           & 0.18                                    & 0.075                           & 0.058                           & 0.149                                  & 0.126         \\ \hline
\end{tabular}%
}
\caption{The p-values obtained from paired t-tests for Verbal Confidence, Accuracy, and ECE between the No Evidence baseline and other types of evidence.}
\label{tab:pvalue_verbal}
\end{table*}
\begin{table*}[!hbt]
\resizebox{\textwidth}{!}{%
\begin{tabular}{l|cccccc}
\hline
           & \multicolumn{1}{l}{\textbf{NO\_EVI-EVI}} & \multicolumn{1}{l}{\textbf{NO\_EVI-Coincidence}} & \multicolumn{1}{l}{\textbf{NO\_EVI-IRR}} & \multicolumn{1}{l}{\textbf{NO\_EVI-Conflict}} & \multicolumn{1}{l}{\textbf{NO\_EVI-Incomplete}} & \multicolumn{1}{l}{\textbf{NO\_EVI-Contradiction}} \\ \hline
Confidence & 0.062                            & 0.074                                  & 0.012                           & 0.445                           & 0.082                                  & 0.056                                     \\
ACC        & 0.082                           & 0.057                                   & 0.052                           & 0.001                           & 0.09                                  & 0.073                                     \\
ECE        & 0.079                           & 0.068                                    & 0.218                           & 0.006                           & 0.096                                  & 0.073         \\ \hline
\end{tabular}%
}
\caption{The p-values obtained from paired t-tests for Token Probability Confidence, Accuracy, and ECE between the No Evidence baseline and other types of evidence.}
\label{tab:pvalue_token}
\end{table*}
\begin{table*}[!hbt]
\resizebox{\textwidth}{!}{%
\begin{tabular}{l|cccccc}
\hline
           & \multicolumn{1}{l}{\textbf{NO\_EVI-EVI}} & \multicolumn{1}{l}{\textbf{NO\_EVI-Coincidence}} & \multicolumn{1}{l}{\textbf{NO\_EVI-IRR}} & \multicolumn{1}{l}{\textbf{NO\_EVI-Conflict}} & \multicolumn{1}{l}{\textbf{NO\_EVI-Incomplete}} & \multicolumn{1}{l}{\textbf{NO\_EVI-Contradiction}} \\ \hline
Confidence & 0.069                            & 0.083                                  & 0.366                           & 0.392                           & 0.085                                  & 0.06                                     \\
ACC        & 0.073                           & 0.048                                   & 0.07                           & 0.015                           & 0.105                                  & 0.062                                     \\
ECE        & 0.037                           & 0.016                                    & 0.248                           & 0.181                           & 0.059                                  & 0.04         \\ \hline
\end{tabular}%
}
\caption{The p-values obtained from paired t-tests for Sampling method, Accuracy, and ECE between the No Evidence baseline and other types of evidence.}
\label{tab:pvalue_sample}
\end{table*}

\clearpage
\bigbreak

\section{Results of Strength of evidence task}\label{apdx:strength}

\begin{figure*}[hbt!]

  \centering

  \includegraphics[width=\textwidth]{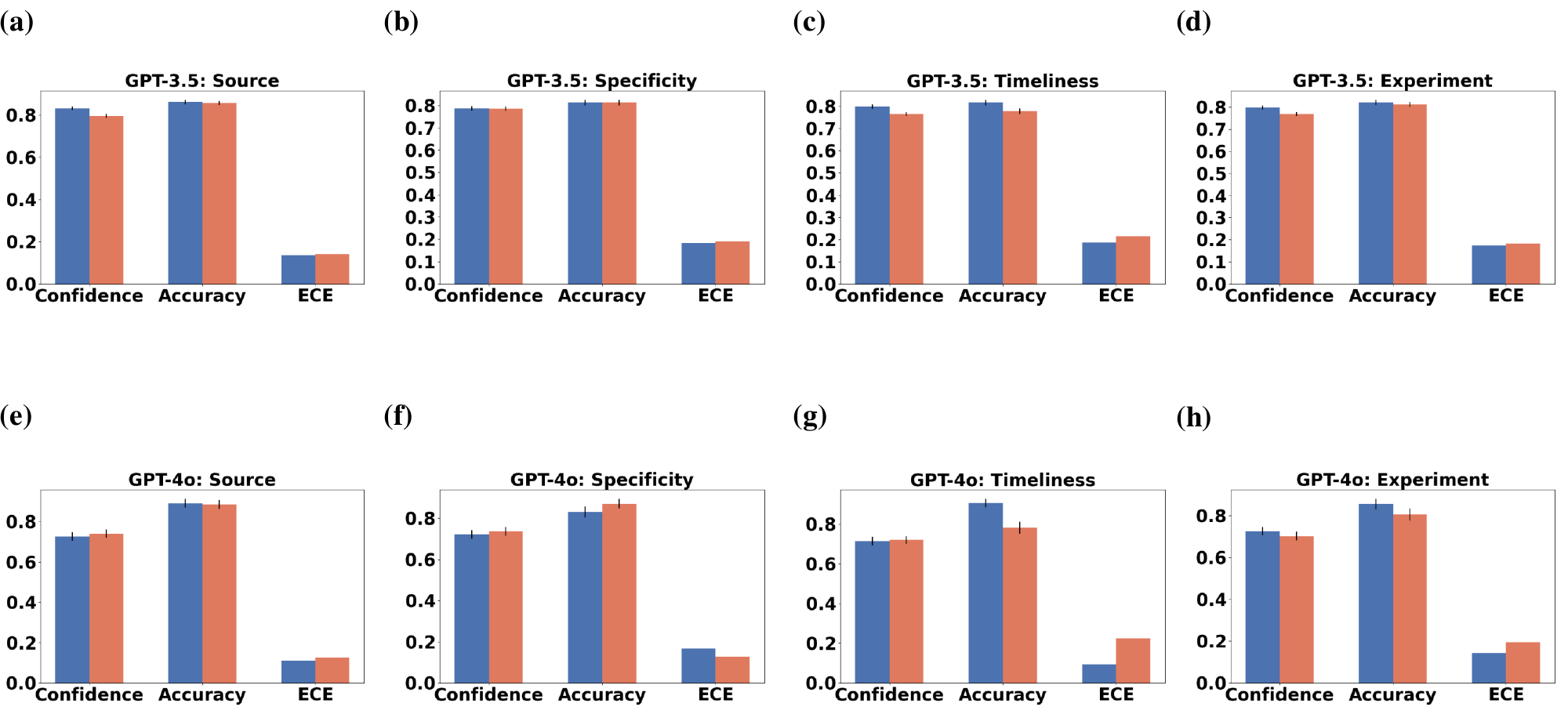}

 \captionsetup{}

  \caption{The results of the Strength of Evidence task on the SciQ dataset with token probability method. The blue bar represents the cases where the strength of evidence is high. Specifically, the blue bar indicates the context from more credible sources, more specific, recent, and experimental evidence, while the red color represents less credible sources, less specific, old, and observational evidence.
}

  \label{fig:strength_token}

\end{figure*}

\begin{figure*}[hbt!]

  \centering

  \includegraphics[width=\textwidth]{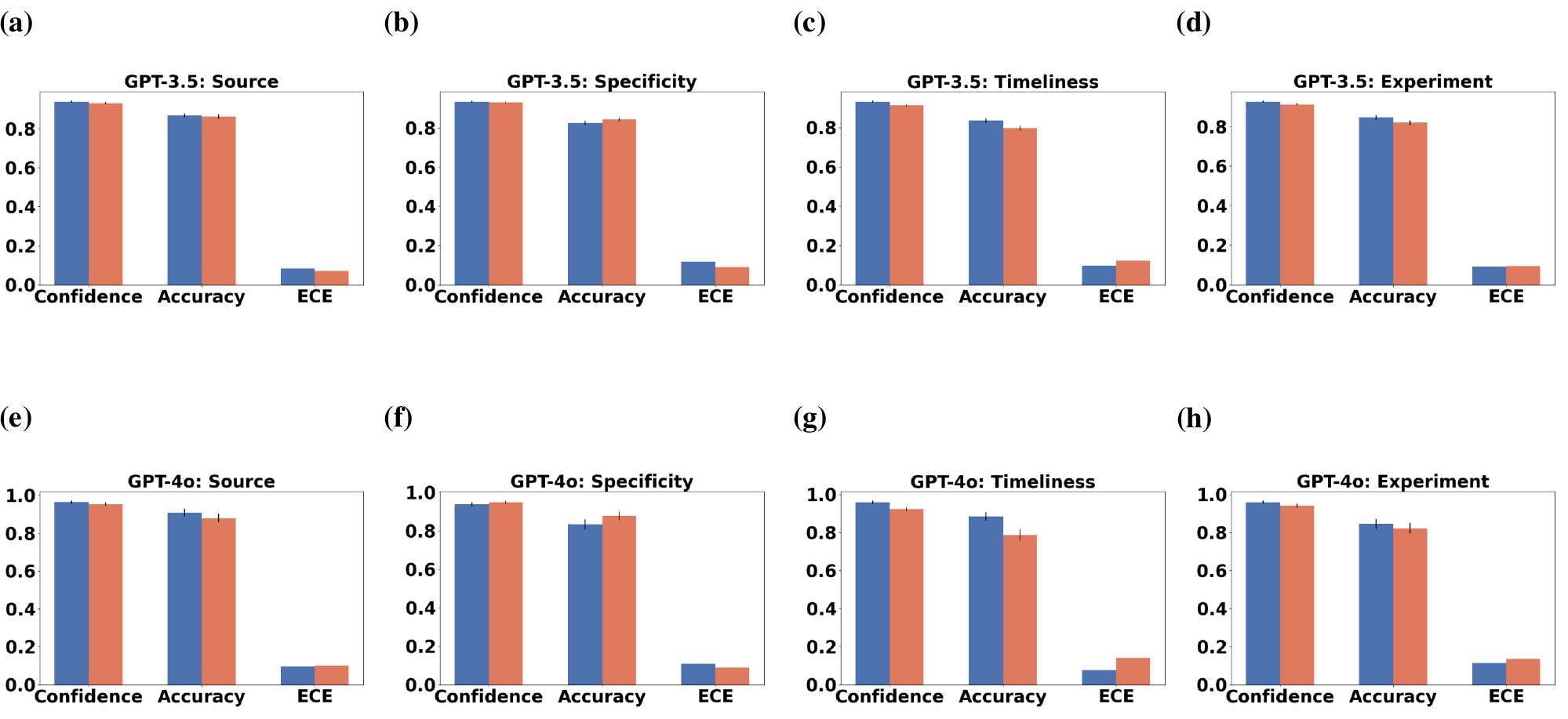}

 \captionsetup{}

  \caption{The results of the Strength of Evidence task on the SciQ dataset with sampling method. The blue bar represents the cases where the strength of evidence is high. Specifically, the blue bar indicates the context from more credible sources, more specific, recent, and experimental evidence, while the red color represents less credible sources, less specific, old, and observational evidence.
}

  \label{fig:strength_sampling}

\end{figure*}

\clearpage
\bigbreak

\section{Results of p\-value for  Strength of evidence task}\label{apdx:pvalue_strength}
\begin{table*}[!hbt]
\resizebox{\textwidth}{!}{%
\begin{tabular}{l|p{0.2\textwidth}p{0.2\textwidth}p{0.2\textwidth}}
\hline
\multicolumn{4}{c}{\textbf{P-values between more reliable evidence and less reliable evidence}} \\ \hline
Method           & \textbf{Verbal} & \textbf{Token} & \textbf{Sampling}  \\ \hline
Confidence & 0.015           & 0.758          & 0.038              \\
ACC        & 0.486           & 0.217          & 0.222              \\
ECE        & 0.401           & 0.048          & 0.817              \\ \hline
\end{tabular}%
}
\caption{The p-values obtained from paired t-tests for confidence, accuracy, and ECE between the less reliable evidence and strong evidence.}
\label{tab:pvalue_strength}
\end{table*}

\clearpage
\bigbreak

\section{Results of Gemini-1.5-flash}\label{apdx:gemini}

\begin{table*}[!hbt]
\resizebox{\textwidth}{!}{%
\begin{tabular}{c|c|cccccccc}
\hline
                               & \textbf{Dataset}     &\textbf{Metric}   & \textbf{No\_EVI} & \textbf{EVI}   & \textbf{Coincidence} & \textbf{Irrelevant} & \textbf{Conflict} & \textbf{Incomplete} & \textbf{Contradiction} \\ \hline \hline
\multirow{9}{*}{\textbf{Gemini-1.5-Flash}} & \multirow{3}{*}{SciQ}  & Confidence  & 0.93   & 0.983 & 0.851       & 0.554      & 0.771    & 0.936      & 0.96         \\
                               &                         & Accuracy \(\uparrow\)  & 0.65   & 0.86  & 0.885       & 0.66      & 0.62      & 0.79      & 0.87         \\
                               &                         & ECE  \(\downarrow\)    & 0.28   & 0.123 & 0.082       & 0.263      & 0.218    & 0.146      & 0.09          \\     \cline{2-10} 
                               & \multirow{3}{*}{Trivia} & Confidence             & 0.909   & 0.881  & 0.791    & 0.45      & 0.644    & 0.855      & 0.854         \\
                               &                         & Accuracy \(\uparrow\)  & 0.855   & 0.91 & 0.97       & 0.62      & 0.73    & 0.88      & 0.93         \\
                               &                         & ECE  \(\downarrow\)    & 0.087   & 0.063 & 0.179       & 0.252      & 0.234     & 0.086      & 0.076         \\ \cline{2-10} 
                               & \multirow{3}{*}{GSM8K}  & Confidence             & 0.987   & 1.0   & 0.703    & 0.546      & 0.552    & 0.967       & 0.967         \\
                               &                         & Accuracy \(\uparrow\)  & 0.165    & 0.97 & 0.64       & 0.15      & 0.07    & 0.74      & 0.965         \\
                               &                         & ECE   \(\downarrow\)   & 0.822   & 0.03 & 0.373       & 0.476      & 0.542     & 0.277      & 0.044         \\  \hline

\end{tabular}%

}
\caption{The result of confirmation task with verbal method. We used 200 samples for Gemini-1.5-Flash due to the cost limit.}
\label{tab:Gemini_main_verbal}
\end{table*}

\begin{table*}[!hbt]
\resizebox{\textwidth}{!}{%
\begin{tabular}{c|c|cccccccc}
\hline
                               & \textbf{Dataset}     &\textbf{Metric}   & \textbf{No\_EVI} & \textbf{EVI}   & \textbf{Coincidence} & \textbf{Irrelevant} & \textbf{Conflict} & \textbf{Incomplete} & \textbf{Contradiction} \\ \hline \hline
\multirow{9}{*}{\textbf{Gemini-1.5-Flash}} & \multirow{3}{*}{SciQ}  & Confidence  & 0.9     & 0.965 & 0.954       & 0.864      & 0.907    & 0.942      & 0.963         \\
                               &                         & Accuracy \(\uparrow\)  & 0.665   & 0.895 & 0.91       & 0.57     & 0.58     & 0.795      & 0.89         \\
                               &                         & ECE  \(\downarrow\)    & 0.335   & 0.105 & 0.09       & 0.43      & 0.42    & 0.205      & 0.11          \\     \cline{2-10} 
                               & \multirow{3}{*}{Trivia} & Confidence             & 0.882   & 0.964 & 0.971     & 0.882      & 0.915    & 0.95      & 0.96         \\
                               &                         & Accuracy \(\uparrow\)  & 0.838   & 0.905 & 0.97       & 0.593      & 0.65    & 0.839      & 0.9         \\
                               &                         & ECE  \(\downarrow\)    & 0.162   & 0.095 & 0.03       & 0.407      & 0.35     & 0.161      & 0.1         \\ \cline{2-10} 
                               & \multirow{3}{*}{GSM8K}  & Confidence             & 0.826   & 0.99  & 0.962    & 0.818      & 0.907    & 0.954       & 0.989         \\
                               &                         & Accuracy \(\uparrow\)  & 0.19    & 0.97 & 0.64       & 0.14      & 0.055    & 0.715      & 0.965         \\
                               &                         & ECE   \(\downarrow\)   & 0.81   & 0.03  & 0.356       & 0.859      & 0.945     & 0.285      & 0.035         \\  \hline

\end{tabular}%

}
\caption{The result of confirmation task with token probability method. We used 200 samples for Gemini-1.5-Flash due to the cost limit.}
\label{tab:Gemini_main_token}
\end{table*}
\begin{table*}[!hbt]
\resizebox{\textwidth}{!}{%
\begin{tabular}{c|c|cccccccc}
\hline
                               & \textbf{Dataset}     &\textbf{Metric}   & \textbf{No\_EVI} & \textbf{EVI}   & \textbf{Coincidence} & \textbf{Irrelevant} & \textbf{Conflict} & \textbf{Incomplete} & \textbf{Contradiction} \\ \hline \hline
\multirow{9}{*}{\textbf{Gemini-1.5-Flash}} & \multirow{3}{*}{SciQ}  & Confidence  & 0.952   & 0.991 & 0.987       & 0.91      & 0.955    & 0.98      & 0.992         \\
                               &                         & Accuracy \(\uparrow\)  & 0.656   & 0.898 & 0.911       & 0.57      & 0.617      & 0.799      & 0.904         \\
                               &                         & ECE  \(\downarrow\)    & 0.294   & 0.107 & 0.081       & 0.339      & 0.337    & 0.186      & 0.09          \\     \cline{2-10} 
                               & \multirow{3}{*}{Trivia} & Confidence             & 0.929   & 0.969  & 0.929    & 0.885      & 0.907    & 0.969      & 0.99         \\
                               &                         & Accuracy \(\uparrow\)  & 0.839   & 0.894 & 0.839      & 0.599      & 0.658    & 0.84      & 0.894         \\
                               &                         & ECE  \(\downarrow\)    & 0.11    & 0.076 & 0.11       & 0.293      & 0.248     & 0.134      & 0.096         \\ \cline{2-10} 
                               & \multirow{3}{*}{GSM8K}  & Confidence             & 0.72    & 0.996  & 0.947    & 0.705     & 0.852    & 0.936       & 0.994         \\
                               &                         & Accuracy \(\uparrow\)  & 0.19    & 0.97 & 0.633      & 0.151     & 0.061    & 0.721      & 0.97         \\
                               &                         & ECE   \(\downarrow\)   & 0.53   & 0.034 & 0.314       & 0.554      & 0.791     & 0.215      & 0.032         \\  \hline

\end{tabular}%

}
\caption{The result of confirmation task with sampling method. We used 200 samples for Gemini-1.5-Flash due to the cost limit.}
\label{tab:Gemini_main_sampling}
\end{table*}

\begin{table*}[!hbt]
\resizebox{\textwidth}{!}{%
\begin{tabular}{c|c|ccccccccc}
\hline
                               & \textbf{Dataset}     &\textbf{Metric}   & \textbf{High Source} & \textbf{Low Source}   & \textbf{High Spec} & \textbf{Low Spec} & \textbf{Recent} & \textbf{Old} & \textbf{Experiment} & \textbf{Observation} \\ \hline \hline
\multirow{3}{*}{\textbf{Gemini-1.5-Flash}} & \multirow{3}{*}{SciQ}  & Confidence  & 0.981   & 0.822 & 0.976       & 0.944      & 0.965    & 0.917      & 0.974   &0.933      \\
                               &                         & Accuracy \(\uparrow\)  & 0.865   & 0.855 & 0.795       & 0.87       & 0.86     & 0.76       & 0.83     &0.785     \\
                               &                         & ECE  \(\downarrow\)    & 0.119   & 0.065 & 0.181       & 0.084      & 0.105    & 0.168      & 0.158   &0.148         \\  \hline

\end{tabular}%

}
\caption{The result of the strength of evidence task with the verbal method. We used 200 samples for Gemini-1.5-Flash due to the cost limit.}
\label{tab:Gemini_str_verbal}
\end{table*}

\begin{table*}[!hbt]
\resizebox{\textwidth}{!}{%
\begin{tabular}{c|c|ccccccccc}
\hline
                               & \textbf{Dataset}     &\textbf{Metric}   & \textbf{High Source} & \textbf{Low Source}   & \textbf{High Spec} & \textbf{Low Spec} & \textbf{Recent} & \textbf{Old} & \textbf{Experiment} & \textbf{Observation} \\ \hline \hline
\multirow{3}{*}{\textbf{Gemini-1.5-Flash}} & \multirow{3}{*}{SciQ}  & Confidence  & 0.944   & 0.955 & 0.947     & 0.942     & 0.943    & 0.936      & 0.942   &0.955      \\
                               &                         & Accuracy \(\uparrow\)  & 0.855   & 0.855 & 0.8       & 0.85      & 0.88     & 0.76       & 0.835     &0.81     \\
                               &                         & ECE  \(\downarrow\)    & 0.145   & 0.145 & 0.2       & 0.15      & 0.12    & 0.24      & 0.165   &0.19          \\  \hline

\end{tabular}%

}
\caption{The result of the strength of evidence task with the token probability method. We used 200 samples for Gemini-1.5-Flash due to the cost limit.}
\label{tab:Gemini_str_token}
\end{table*}

\begin{table*}[!hbt]
\resizebox{\textwidth}{!}{%
\begin{tabular}{c|c|ccccccccc}
\hline
                               & \textbf{Dataset}     &\textbf{Metric}   & \textbf{High Source} & \textbf{Low Source}   & \textbf{High Spec} & \textbf{Low Spec} & \textbf{Recent} & \textbf{Old} & \textbf{Experiment} & \textbf{Observation} \\ \hline \hline
\multirow{3}{*}{\textbf{Gemini-1.5-Flash}} & \multirow{3}{*}{SciQ}  & Confidence  & 0.974   & 0.986 & 0.978       & 0.981      & 0.97     & 0.964      & 0.973   &0.982      \\
                               &                         & Accuracy \(\uparrow\)  & 0.879   & 0.854 & 0.809       & 0.853      & 0.879     & 0.774       & 0.83     &0.791     \\
                               &                         & ECE  \(\downarrow\)    & 0.106   & 0.133 & 0.169       & 0.128      & 0.091    & 0.193      & 0.15   &0.193           \\  \hline

\end{tabular}%

}
\caption{The result of the strength of evidence task with the sampling method. We used 200 samples for Gemini-1.5-Flash due to the cost limit.}
\label{tab:Gemini_str_samplin}
\end{table*}

\clearpage
\bigbreak

\section{Results of Ablation study on the ratio of golden evidence}\label{apdx:golden}
\begin{table*}[!hbt]
\resizebox{\textwidth}{!}{%
\begin{tabular}{c|c|cccccccccccc}
\hline
                               & \textbf{Dataset}     &\textbf{Metric}   &\textbf{Conflict\_30} &\textbf{Conflict\_50}   & \textbf{Conflict\_80} &\textbf{Conflict\_100} &\textbf{Incomplete\_30} &\textbf{Incomplete\_50} &\textbf{Incomplete\_80} &\textbf{Contradict\_30} &\textbf{Contradict\_50}   &\textbf{Contradict\_80} &\textbf{Contradict\_100} \\ \hline \hline
\multirow{9}{*}{\textbf{GPT-3.5-turbo}} & \multirow{3}{*}{SciQ}  & Confidence     & 0.932   & 0.912 & 0.88        & 0.827      & 0.935    & 0.928     & 0.906  &0.947 &0.945 &0.943 &0.95          \\
                               &                         & Accuracy \(\uparrow\)  & 0.803    & 0.744 & 0.745       & 0.572      & 0.791  & 0.77       & 0.693   &0.827 &0.847 &0.833 &0.843       \\
                               &                         & ECE  \(\downarrow\)    & 0.138    & 0.184 & 0.216       & 0.304     & 0.152   & 0.161      & 0.216   &0.127 &0.108 &0.122 &0.115      \\ \cline{2-14} 
                               
                               & \multirow{3}{*}{Trivia} & Confidence             & 0.908   & 0.887 & 0.851       & 0.797      & 0.909    & 0.897      &0.872  &0.922 &0.925 &0.923 &0.925       \\                            
                               &                         & Accuracy \(\uparrow\)  & 0.859   & 0.843 & 0.785       & 0.702      & 0.867    & 0.86       &0.839  &0.874 &0.869 &0.857 &0.864        \\
                               &                         & ECE  \(\downarrow\)    & 0.072   & 0.087 & 0.136       & 0.211      & 0.049    & 0.058      &0.07   &0.07  &0.076 &0.09  &0.085       \\ \cline{2-14} 
                               
                               & \multirow{3}{*}{GSM8K}  & Confidence             & 0.961   & 0.956 & 0.949       & 0.931      & 0.98     & 0.96       & 0.938  &0.95  &0.949 &0.959 &0.974      \\                               
                               &                         & Accuracy \(\uparrow\)  & 0.772   & 0.5   & 0.267       & 0.023      & 0.853    & 0.666      & 0.361  &0.796 &0.777 &0.791 &0.761       \\
                               &                         & ECE  \(\downarrow\)    & 0.203   & 0.466 & 0.685       & 0.912      & 0.135    & 0.307      & 0.578  &0.197 &0.195 &0.197 &0.234      \\ \hline
                               
\multirow{9}{*}{\textbf{GPT-4o}}        & \multirow{3}{*}{SciQ}   & Confidence     & 0.967   & 0.93  & 0.9         & 0.875      & 0.969   & 0.948      & 0.909  &0.98  &0.977 &0.968 &0.963       \\
                               &                         & Accuracy \(\uparrow\)   & 0.88    & 0.839 & 0.734       & 0.675      & 0.87    & 0.82       & 0.764  &0.904 &0.905 &0.92  &0.92          \\
                               &                         & ECE   \(\downarrow\)    & 0.087   & 0.101 & 0.166       & 0.2        & 0.105   & 0.128      & 0.145  &0.082 &0.072 &0.062 &0.058       \\ \cline {2-14} 
                               
                               & \multirow{3}{*}{Trivia} & Confidence             & 0.919   & 0.891 & 0.884       & 0.866      & 0.927    & 0.909      & 0.882  &0.934 &0.927 &0.925 &0.925  \\                             
                               &                         & Accuracy \(\uparrow\)  & 0.96    & 0.92  & 0.915       & 0.86      & 0.96     & 0.945      & 0.925  &0.945 &0.955 &0.96  &0.944        \\
                               &                         & ECE  \(\downarrow\)    & 0.041   & 0.035 & 0.032       & 0.048      & 0.035    & 0.036      & 0.048  &0.021 &0.037 &0.035 &0.039        \\ \cline{2-14} 
                               
                               & \multirow{3}{*}{GSM8K}  & Confidence             & 0.87    & 0.855 & 0.852       & 0.882      & 0.982    & 0.96       & 0.964   &0.971 &0.957 &0.952 &0.951        \\
                               &                         & Accuracy \(\uparrow\)  & 0.795   & 0.64  & 0.27        & 0.165      & 0.935    & 0.774      & 0.585   &0.94  &0.96  &0.97  &0.935       \\
                               &                         & ECE   \(\downarrow\)   & 0.189   & 0.318 & 0.648       & 0.718      & 0.065    & 0.186      & 0.379   &0.031 &0.013 &0.018 &0.026       \\ \hline
\end{tabular}%
}

\caption{The result of the ratio of golden sentence ablation study with verbalized method. We used 200 samples for GPT-4o due to the cost limit. We modified the number of negated sentences, the number of sentences in incomplete evidence, and the number of contradictory sentences in contradictory evidence and measured Confidence, Accuracy, and ECE. For example, Conflict\_80 means 80\% of the entire sentences have been replaced into conflicting sentences, and Incomplete\_80 means 80\% of sentences have been deleted. Additionally, Contradict\_80 refers 80\% of evidence has been negated and appended to the evidence.}
\label{tab:ablation_table_verbal}
\end{table*}
\begin{table*}[!hbt]
\resizebox{\textwidth}{!}{%
\begin{tabular}{c|c|cccccccccccc}
\hline
                               & \textbf{Dataset}     &\textbf{Metric}   &\textbf{Conflict\_30} &\textbf{Conflict\_50}   & \textbf{Conflict\_80} &\textbf{Conflict\_100} &\textbf{Incomplete\_30} &\textbf{Incomplete\_50} &\textbf{Incomplete\_80} &\textbf{Contradict\_30} &\textbf{Contradict\_50}   &\textbf{Contradict\_80} &\textbf{Contradict\_100} \\ \hline \hline
\multirow{9}{*}{\textbf{GPT-3.5-turbo}} & \multirow{3}{*}{SciQ}  & Confidence     & 0.745    & 0.725 & 0.677       & 0.638      & 0.751   & 0.723      & 0.684   &0.764 &0.764 &0.765 &0.76          \\
                               &                         & Accuracy \(\uparrow\)  & 0.785    & 0.746 & 0.693       & 0.6        & 0.792   & 0.741      & 0.68    &0.831 &0.837 &0.85  &0.846       \\
                               &                         & ECE  \(\downarrow\)    & 0.2      & 0.238 & 0.297       & 0.381      & 0.205   & 0.245      & 0.308   &0.164 &0.16  &0.152 &0.151   \\ \cline{2-14} 
                               
                               & \multirow{3}{*}{Trivia} & Confidence             & 0.854   & 0.831 & 0.8         & 0.759      & 0.853    & 0.843      &0.822  &0.878 &0.849 &0.851 &0.857       \\                            
                               &                         & Accuracy \(\uparrow\)  & 0.863   & 0.808 & 0.742       & 0.668      & 0.851    & 0.852      &0.814  &0.873 &0.857 &0.867 &0.851        \\
                               &                         & ECE  \(\downarrow\)    & 0.2     & 0.187 & 0.251       & 0.326      & 0.146    & 0.141      &0.178  &0.132 &0.139 &0.136 &0.147       \\ \cline{2-14} 
                               
                               & \multirow{3}{*}{GSM8K}  & Confidence             & 0.877   & 0.807 & 0.765       & 0.738      & 0.894    & 0.765      & 0.532  &0.842 &0.801 &0.796 &0.801      \\                               
                               &                         & Accuracy \(\uparrow\)  & 0.803   & 0.518 & 0.262       & 0.028      & 0.881    & 0.677      & 0.384  &0.825 &0.775 &0.777 &0.741       \\
                               &                         & ECE  \(\downarrow\)    & 0.207   & 0.469 & 0.725       & 0.939      & 0.118    & 0.299      & 0.534  &0.172 &0.222 &0.211 &0.257      \\ \hline
                               
\multirow{9}{*}{\textbf{GPT-4o}}        & \multirow{3}{*}{SciQ}   & Confidence     & 0.778   & 0.751 & 0.712       & 0.653      & 0.785   & 0.744      & 0.669  &0.822 &0.813 &0.824 &0.828       \\
                               &                         & Accuracy \(\uparrow\)   & 0.885   & 0.84  & 0.78        & 0.655      & 0.88    & 0.835      & 0.775  &0.925 &0.925 &0.925 &0.92          \\
                               &                         & ECE   \(\downarrow\)    & 0.116   & 0.169 & 0.236       & 0.334      & 0.12    & 0.165      & 0.216  &0.075 &0.078 &0.074 &0.077       \\ \cline {2-14} 
                               
                               & \multirow{3}{*}{Trivia} & Confidence             & 0.905   & 0.85  & 0.853       & 0.824      & 0.911    & 0.889      & 0.858  &0.913 &0.91  &0.914 &0.918  \\                             
                               &                         & Accuracy \(\uparrow\)  & 0.94    & 0.9   & 0.86        & 0.82       & 0.95     & 0.94       & 0.925  &0.96  &0.95  &0.945 &0.944        \\
                               &                         & ECE  \(\downarrow\)    & 0.058   & 0.104 & 0.146       & 0.173      & 0.045    & 0.064      & 0.078  &0.04  &0.05  &0.055 &0.052        \\ \cline{2-14} 
                               
                               & \multirow{3}{*}{GSM8K}  & Confidence             & 0.765   & 0.611  & 0.421       & 0.372      & 0.856    & 0.755      & 0.599   &0.856 &0.842 &0.851 &0.862        \\
                               &                         & Accuracy \(\uparrow\)  & 0.835   & 0.61   & 0.351       & 0.191      & 0.945    & 0.83       & 0.59    &0.95  &0.955 &0.965 &0.955       \\
                               &                         & ECE   \(\downarrow\)   & 0.16    & 0.349  & 0.614       & 0.74       & 0.055    & 0.127      & 0.393   &0.05  &0.037 &0.035 &0.041       \\ \hline
\end{tabular}%
}

\caption{The result of the ratio of golden sentence ablation study with token probability. We used 200 samples for GPT-4o due to the cost limit. We modified the number of negated sentences, the number of sentences in incomplete evidence, and the number of contradictory sentences in contradictory evidence and measured Confidence, Accuracy, and ECE.  For example, Conflict\_80 means 80\% of the entire sentences have been replaced into conflicting sentences, and Incomplete\_80 means 80\% of sentences have been deleted. Additionally, Contradict\_80 refers 80\% of evidence has been negated and appended to the evidence.}
\label{tab:ablation_table_token}
\end{table*}
\begin{table*}[!hbt]
\resizebox{\textwidth}{!}{%
\begin{tabular}{c|c|cccccccccccc}
\hline
                               & \textbf{Dataset}     &\textbf{Metric}   &\textbf{Conflict\_30} &\textbf{Conflict\_50}   & \textbf{Conflict\_80} &\textbf{Conflict\_100} &\textbf{Incomplete\_30} &\textbf{Incomplete\_50} &\textbf{Incomplete\_80} &\textbf{Contradict\_30} &\textbf{Contradict\_50}   &\textbf{Contradict\_80} &\textbf{Contradict\_100} \\ \hline \hline
\multirow{9}{*}{\textbf{GPT-3.5-turbo}} & \multirow{3}{*}{SciQ}  & Confidence     & 0.904    & 0.885 & 0.865       & 0.828      & 0.906   & 0.888      & 0.87    &0.914 &0.922 &0.921 &0.918         \\
                               &                         & Accuracy \(\uparrow\)  & 0.822    & 0.77  & 0.706       & 0.616      & 0.813   & 0.777      & 0.71    &0.859 &0.853 &0.856 &0.852       \\
                               &                         & ECE  \(\downarrow\)    & 0.091    & 0.115 & 0.158       & 0.211      & 0.093   & 0.111      & 0.165   &0.064 &0.074 &0.072 &0.07       \\ \cline{2-14} 
                               
                               & \multirow{3}{*}{Trivia} & Confidence             & 0.927   & 0.917 & 0.885       & 0.862      & 0.929    & 0.924      &0.905  &0.935 &0.934 &0.936 &0.931       \\                            
                               &                         & Accuracy \(\uparrow\)  & 0.864   & 0.829 & 0.776       & 0.693      & 0.866    & 0.856      &0.83   &0.882 &0.884 &0.869 &0.863        \\
                               &                         & ECE  \(\downarrow\)    & 0.069   & 0.093 & 0.129       & 0.17       & 0.067    & 0.072      &0.085  &0.058 &0.076 &0.078 &0.072       \\ \cline{2-14} 
                               
                               & \multirow{3}{*}{GSM8K}  & Confidence             & 0.937   & 0.883 & 0.849       & 0.838      & 0.949    & 0.924      & 0.656  &0.874 &0.848 &0.845 &0.861      \\                               
                               &                         & Accuracy \(\uparrow\)  & 0.805   & 0.531 & 0.267       & 0.028      & 0.896    & 0.856      & 0.417  &0.802 &0.757 &0.736 &0.722       \\
                               &                         & ECE  \(\downarrow\)    & 0.133   & 0.352 & 0.583       & 0.81       & 0.06     & 0.072      & 0.239  &0.079 &0.092 &0.123 &0.152      \\ \hline
                               
\multirow{9}{*}{\textbf{GPT-4o}}        & \multirow{3}{*}{SciQ}   & Confidence     & 0.943   & 0.922 & 0.904       & 0.871      & 0.954   & 0.923      & 0.906  &0.958 &0.965 &0.959 &0.957       \\
                               &                         & Accuracy \(\uparrow\)   & 0.893   & 0.848 & 0.807       & 0.698      & 0.887   & 0.84       & 0.77   &0.929 &0.933 &0.938 &0.934         \\
                               &                         & ECE   \(\downarrow\)    & 0.078   & 0.109 & 0.114       & 0.187      & 0.086   & 0.132      & 0.137  &0.063 &0.066 &0.045 &0.075       \\ \cline {2-14} 
                               
                               & \multirow{3}{*}{Trivia} & Confidence             & 0.969   & 0.959 & 0.942       & 0.918      & 0.97     & 0.966      & 0.954  &0.97  &0.97  &0.965 &0.974  \\                             
                               &                         & Accuracy \(\uparrow\)  & 0.969   & 0.919 & 0.872       & 0.843      & 0.98     & 0.924      & 0.934  &0.949 &0.959 &0.954 &0.974        \\
                               &                         & ECE  \(\downarrow\)    & 0.018   & 0.078 & 0.075       & 0.122      & 0.035    & 0.042      & 0.028  &0.028 &0.038 &0.046 &0.027        \\ \cline{2-14} 
                               
                               & \multirow{3}{*}{GSM8K}  & Confidence             & 0.862   & 0.742 & 0.581       & 0.529      & 0.943    & 0.875      & 0.741   &0.948 &0.957 &0.952 &0.944        \\
                               &                         & Accuracy \(\uparrow\)  & 0.882   & 0.685 & 0.407       & 0.224      & 0.943    & 0.829      & 0.622   &0.964 &0.969 &0.964 &0.954       \\
                               &                         & ECE   \(\downarrow\)   & 0.059   & 0.074 & 0.174       & 0.305      & 0.047    & 0.103      & 0.119   &0.067 &0.051 &0.063 &0.08        \\ \hline
\end{tabular}%
}

\caption{The result of the ratio of golden sentence ablation study with sampling method. We used 200 samples for GPT-4o due to the cost limit. We modified the number of negated sentences, the number of sentences in incomplete evidence, and the number of contradictory sentences in contradictory evidence and measured Confidence, Accuracy,
and ECE. For example, Conflict\_80 means 80\% of the entire sentences have been replaced into conflicting sentences, and Incomplete\_80 means 80\% of sentences have been deleted. Additionally, Contradict\_80 refers 80\% of evidence has been negated and appended to the evidence.}
\label{tab:ablation_table_sampling}
\end{table*}

\clearpage
\bigbreak

\section{Results of Ablation study on irrelevant evidence}\label{apdx:irrelevant}
\begin{figure*}[hbt!]

  \centering

  \includegraphics[width=\textwidth]{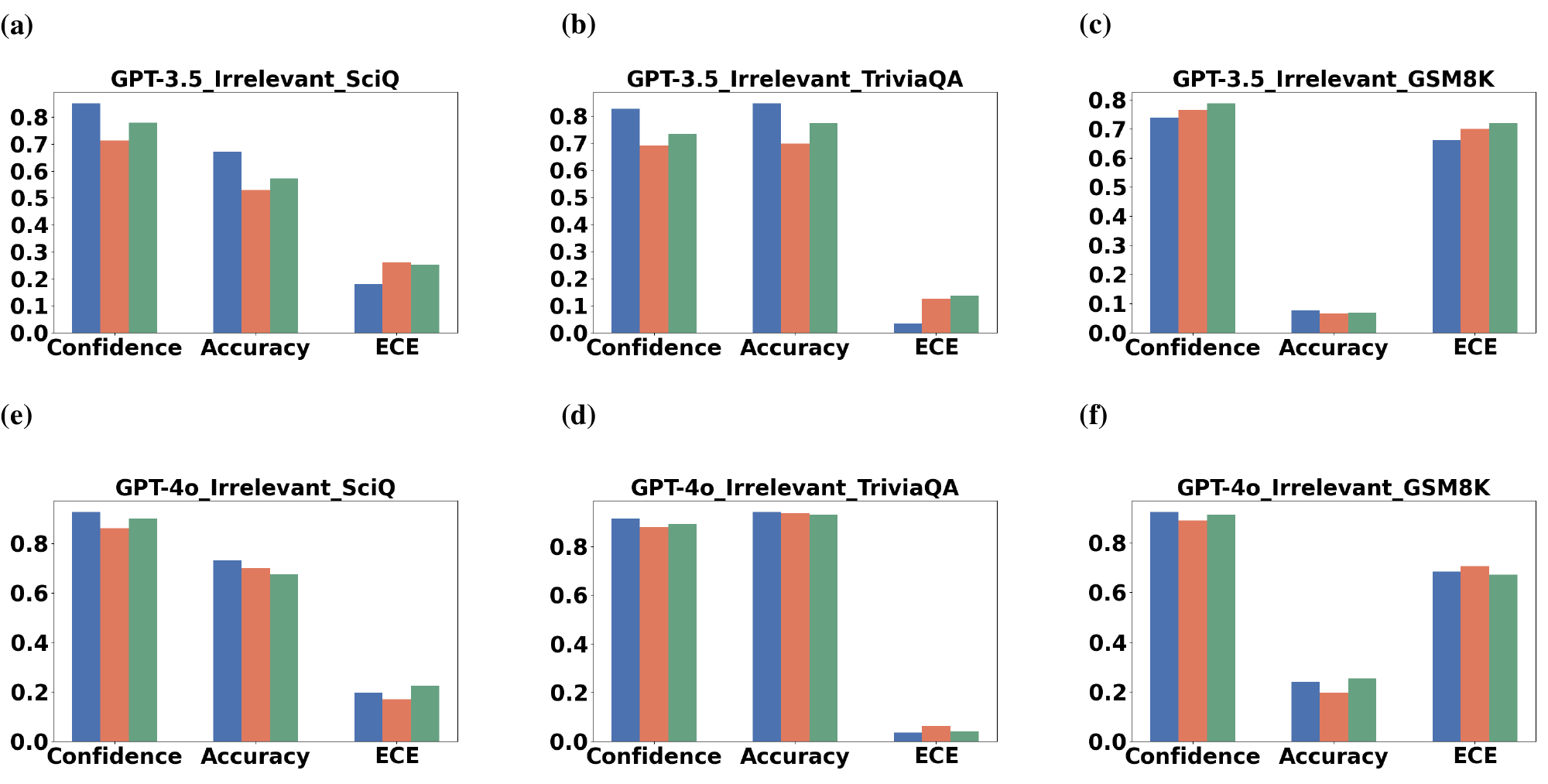}

 \captionsetup{}

  \caption{The results of ablation for irrelevant evidence. The blue bar represents the result of no evidence \(P(H)\), serving as a baseline. The red bar results from irrelevant evidence by replacing evidence from other samples within the same dataset explained in section 3.2. The green bar represents irrelevant evidence from another dataset. 
}

  \label{fig:irrelevant}

\end{figure*}

\clearpage
\bigbreak

\twocolumn
\section{Experimental Detail}\label{apdx:exepriment}

\subsection{Hyperparameter}\label{subsec:hyperparameter}
We utilized OpenAI's API to create a dataset containing evidence and conducted inference experiments. Specifically, we used GPT-4-0613 to generate Negated evidence, Coincidental evidence, and Contradictory evidence required for the confirmation task, and gpt-4o-2024-05-13 to create evidence necessary for the strength of evidence. The inference was performed using GPT-3.5-0125 and GPT-4o-2024-05-13 with settings of temperature=1.0 and top\_p=1.0.

\subsection{Evaluation Detail}\label{subsec:eval}
According to \citep{kuhn2023semantic}, for the SciQ and TriviaQA datasets, we considered a model's response as correct if its Rouge-L score \citep{lin-2004-rouge} with the golden label is 0.3 or higher. For GSM8K, only responses that were an exact match with the golden label were considered correct.

For sampling method for measuring confidence, we set the ratio of most frequent response as the confidence. As the datasets are open-ended question, we should consider the synonym of each responses. In order to handle this, we used GPT-4o-2024-05-13 to capture the semantic similarity and calculate the frequency of the most common response.

For measuring p-value, we conducted two-sided paired t-tests for p-value calculations. Instead of using individual samples, we performed the tests on dataset-level metrics by comparing the mean metrics (e.g., accuracy) under different evidence conditions (e.g., No Evidence vs. Golden Evidence). Specifically, we calculated the p-value for accuracy by comparing six results (three datasets × two models) for No Evidence with the corresponding six results for Golden Evidence.
\subsection{Dataset}\label{subsec:dataset}
For SciQ and GSM8K, we extracted the samples containing the explanation, including more than 4 sentences to create various proportions of negated sentences in the ablation study. Similarly, for trivia QA, we used the explanation\footnote{We used the context of each question as evidence. For the context of each sample, we used the positive passage in https://huggingface.co/datasets/Tevatron/wikipedia-trivia.} including more than 4 sentences and extracting 1000 samples. We generated negated sentences using GPT-4-0613 for negated and contradictory evidence and filtered out samples containing incorrect sentences. Similarly, we used GPT-4o-2024-05-13 for generating Strength of Evidence task and also filtered out the generated strength of evidence that included a wrong template. The total number of samples is shown in Table \ref{tab:sample_num_confirm} and Table \ref{tab:sample_num_strength}. We used all these samples when inferencing with GPT-3.5-turbo and 200 samples for GPT-4o-2024-05-13.

\clearpage
\bigbreak

\onecolumn

\begin{table}[hbt!]
\resizebox{\textwidth}{!}{%
\begin{tabular}{l|lllllll}
         & NO\_EVI & EVI  & Coincidence & Irrelevant & Negation & Incomplete & Contradiction \\ \hline \hline
SciQ     & 1095    & 1095 & 1095        & 1095       & 991      & 1095       & 991           \\
TriviaQA & 1000    & 1000 & 1000        & 1000       & 798      & 1000       & 798           \\
GSM8K    & 622     & 622  & 622         & 622        & 618     & 622        & 618     \\  \hline   
\end{tabular}
}
\caption{The number of samples for the Confirmation task dataset.}\label{tab:sample_num_confirm}
\end{table}

\begin{table}[hbt!]
\resizebox{\textwidth}{!}{%
\begin{tabular}{l|llllllll}
         & High Credible Source & Low Credible Source & High Specificity &Low Specificity & Recent & Old & Experiment  & Observation \\ \hline \hline
SciQ     & 1095                 & 1095                & 1093             & 1093           & 1074   & 1074 & 1094      & 1094        \\ \hline   
\end{tabular}
}
\caption{The number of samples for the Strength of evidence task dataset.}\label{tab:sample_num_strength}
\end{table}

\clearpage
\bigbreak

\section{Prompt}\label{apdx:prompt}
In this section, we will show the prompt for inference,

\subsection{Prompt for Inference}\label{subsec:promptinference}
\begin{table}[htb]
\begin{tcolorbox}[toprule=1mm,colback=gray!10!white,colframe=gray!50,title=Verbal Confidence Prompt, fonttitle=\bfseries]
Provide your best guess and the probability that it is correct (0.0 to 1.0) for the following question based on the evidence. \\
Give ONLY the guess and probability, no other words or explanation.\\ 
For example\\
Guess: <most likely guess, as short as possible; not a complete sentence, just the guess!>\\
Probability: <the probability between 0.0 and 1.0 that your guess is correct based on the given evidence , without any extra commentary whatsoever; just the probability!>\\
\#\#\#\textbf{The question}: \{question\}\\
\#\#\#\textbf{The evidence}: \{evidence\}\\
\end{tcolorbox}
\caption{A prompt for verbal confidence and guess of answer from language models. We follow \citep{tian2023just}.}
\label{tab:t5_1}
\end{table}

\begin{table}[htb]
\begin{tcolorbox}[toprule=1mm,colback=gray!10!white,colframe=gray!50,title= Prompt for Token probability and Sampling, fonttitle=\bfseries]
Provide your best guess for the following question based on the evidence. \\
Give ONLY the guess, no other words or explanation. \\
For example\\
Guess: <most likely guess, as short as possible; not a complete sentence, just the guess!>\\
\#\#\#\textbf{The question}: \{question\} \\
\#\#\#\textbf{The evidence}: \{evidence\} \\
\end{tcolorbox}
\caption{A prompt for Token probability and guess of answer from language models. We do not need to extract the confidence by prompt, so all we need is to extract the guess.}
\label{tab:t5_1}
\end{table}

\subsection{Prompt for Generating Evidence}\label{subsec:promptevidence}
\begin{table}[hbt]
\renewcommand{\baselinestretch}{0.93}\selectfont
\begin{tcolorbox}[ toprule=1mm,colback=gray!10!white,colframe=gray!50,title= Prompt for Negating the evidence, fonttitle=\bfseries]
\#\#\#\textbf{Example}: "Biochemical reactions of metabolism can be divided into two general categories: catabolic reactions and anabolic reactions. You can watch an animation showing how the two categories of reactions are related at this URL: http://classes. midlandstech. edu/carterp/courses/bio225/chap05/lecture1. htm." \\
Revise or negate each sentence in the \#\#\#Example with incorrect information yet relevant information. The response \#\#\#Negation should have same number of sentence with \#\#\#Example.\\
\#\#\#\textbf{Negation}:"Biochemical reactions of metabolism are typically classified into only one category: equilibrium reactions. You can view a static image illustrating the isolated function of equilibrium reactions at this URL:http://classes.midlandstech.edu/carterp/courses/bio225/chap05/lecture2.htm."
\\
\#\#\#\textbf{Example}: "An anaerobic organism is any organism that does not need oxygen for growth and even dies in its presence. Obligate anaerobes will die when exposed to atmospheric levels of oxygen. Clostridium perfringens bacteria, which are commonly found in soil around the world, are obligate anaerobes. Infection of a wound by C. perfringens bacteria causes the disease gas gangrene. Obligate anaerobes use molecules other than oxygen as terminal electron acceptors." \\
Revise or negate each sentence in the \#\#\#Example with incorrect information yet relevant information. The response \#\#\#Negation should have same number of sentence with \#\#\#Example.\\
\#\#\#\textbf{Negation}: "An anaerobic organism is any organism that requires oxygen for growth and thrives in its presence. Obligate aerobes will perish when deprived of atmospheric oxygen levels. Staphylococcus aureus bacteria, which are rarely found in aquatic environments, are obligate aerobes. Infection of a wound by S. aureus bacteria causes the disease known as athlete's foot. Obligate aerobes use molecules such as hydrogen or sulfur as terminal electron acceptors."
\\
\#\#\#\textbf{Example}: "The energy of a mechanical wave can travel only through matter. The matter through which the wave travels is called the medium ( plural , media). The medium in the water wave pictured above is water, a liquid. But the medium of a mechanical wave can be any state of matter, even a solid.” \\
Revise or negate each sentence in the \#\#\#Example with incorrect information yet relevant information. The response \#\#\#Negation should have same number of sentence with \#\#\#Example.\\
\#\#\#\textbf{Negation}: "The energy of a mechanical wave can travel through both matter and vacuum. The space through which the wave travels is termed the conduit. The conduit in the water wave pictured above is air, a gas. However, the conduit of a mechanical wave can be exclusively in a gaseous state, not a solid or liquid.”
\\
\#\#\#\textbf{Example}: "What group of animals begins its life in the water, but then spends most of its life on land? Amphibians! Amphibians are a group of vertebrates that has adapted to live in both water and on land. Amphibian larvae are born and live in water, and they breathe using gills. The adults live on land for part of the time and breathe both through their skin and with their lungs as their lungs are not sufficient to provide the necessary amount of oxygen." \\
Revise or negate each sentence in the \#\#\#Example with incorrect information yet relevant information. The response \#\#\#Negation should have same number of sentence with \#\#\#Example.\\
\#\#\#\textbf{Negation}: "What group of animals begins its life on land, but then spends most of its life in water? Reptiles! Reptiles are a group of vertebrates that has adapted to live mainly on land but also in water. Reptile eggs are laid and hatch on land, and they breathe using lungs from birth. The adults live in water for part of the time and breathe exclusively through their lungs as their skin is not permeable enough to facilitate breathing."
\\
\#\#\#\textbf{Example}:\{source\}\\
Revise or negate each sentence in the \#\#\#Example with incorrect information yet relevant information. The response \#\#\#Negation should have same number of sentence with \#\#\#Example. \\
\#\#\#\textbf{Negation}:
\end{tcolorbox}
\caption{A prompt for negating the each sentence in golden evidence.}
\label{tab:t5_1}
\end{table}

\begin{table}[htb]
\begin{tcolorbox}[toprule=1mm,colback=gray!10!white,colframe=gray!50,title= Prompt for Token probability and Sampling, fonttitle=\bfseries]
\#\#\#\textbf{Question}: "What does the pull of the moon's gravity on earth cause?”, Answer: tides \\
Make explanation that realize the answer to the question by knowing from unreasonable guess. You should not express that answer from unreasonable guess is correct for the question.\\
\#\#\#\textbf{Explanation}: "Interestingly, while reviewing some casual photographs taken during a lunar eclipse, a curious pattern emerged that reminded me of a physics class discussion. The photos captured how remarkably calm and aligned everything seemed under the moonlight, which brought to mind the effects of the moon's gravity on Earth. It dawned on me, almost by happenstance, that the moon’s gravitational pull is what primarily drives the ocean tides. This realization was unexpected, as the focus of the eclipse viewing was purely for its beauty and not the scientific implications. It's fascinating how a moment of leisure led to a reminder that while the moon’s gravity tugs at the land, only the water, being fluid, visibly responds, creating the tides we observe."\\

\#\#\#\textbf{Question}: "What living species of bird is considered the largest of the raptors?” Answer: golden eagles
Make explanation that realize the answer to the question by knowing from unreasonable guess. You should not express that answer from unreasonable guess is correct for the question.\\
\#\#\#\textbf{Explanation}: "During a casual discussion at a local café in Northern California, after a day of hiking, I was sharing stories and showing photos I had taken of a large bird, Golden Eagle, I encountered. Not being an expert, I half-jokingly guessed it might be one of the larger bird species due to its impressive size and wingspan."\\

\#\#\#\textbf{Question}: "What is another term for blood clotting?" Answer: coagulation\\
Make explanation that realize the answer to the question by knowing from unreasonable guess. You should not express that answer from unreasonable guess is correct for the question.\\
\#\#\#\textbf{Explanation}:”At a community health fair, while casually discussing various health topics, an attendee mentioned his surprise at how quickly a minor cut he had stopped bleeding. Not being a healthcare professional, I playfully suggested it might be due to his body being exceptionally good at 'coagulating'—a term I recalled from a movie rather than a biology class.”\\

\#\#\#\textbf{Question}: "What is the protein in red blood cells which transports oxygen around the body?”, Answer: hemoglobin\\
Make explanation that realize the answer to the question by knowing from unreasonable guess. You should not express that answer from unreasonable guess is correct for the question.\\
\#\#\#\textbf{Explanation}: “During a spirited trivia night with friends, where various odd facts were thrown around, someone challenged the group to name the protein in red blood cells responsible for transporting oxygen. Without any medical background, I haphazardly threw out 'hemoglobin' as a guess, mainly because it sounded scientific and related to blood. My guess was influenced by overhearing a conversation earlier that day at the gym where someone mentioned they were taking iron supplements to improve their 'hemoglobin levels' for better endurance”\\

\#\#\#\textbf{Question}: \{question\}, \textbf{Answer}: \{answer\}\\
Make explanation that realize the answer to the question by knowing from unreasonable guess. You should not express that answer from unreasonable guess is correct for the question.\\
\#\#\#\textbf{Explanation}:

\end{tcolorbox}
\caption{A prompt for making coincidental evidence.}
\label{tab:t5_1}
\end{table}

\begin{table}[htb]
\begin{tcolorbox}[toprule=1mm,colback=gray!10!white,colframe=gray!50,title= Prompt for Making credible evidence, fonttitle=\bfseries]
\#\#\#\textbf{question}: "What substance does the phillosopher stone change the base material to?" \\
\#\#\#\textbf{answer}:  "gold"\\

For this \#\#\#question, \#\#\#answer pairs, make 3 evidences with difference power of evidence in the aspect of Source Credibility.\\

\#\#\#\textbf{Highly Credible Source}: "A leading professor of alchemy at a renowned university published a peer-reviewed paper documenting the transmutation of lead into gold using the Philosopher's Stone."\\

\#\#\#\textbf{Moderately Credible Source}: "A respected independent alchemist reported successful transmutations in his personal journal."\\

\#\#\#\textbf{Low Credibility Source}: "An anonymous blog post claims to have discovered the Philosopher's Stone and successfully converted lead into gold."\\

\#\#\#\textbf{question}: "Compounds with aluminum and silicon are commonly found in the clay fractions of soils derived from what?"\\
\#\#\#\textbf{answer}: "volcanic ash"\\

For this \#\#\#question, \#\#\#answer pairs, make 3 evidences with difference power of evidence in the aspect of Source Credibility.\\

\#\#\#\textbf{Highly Credible Source}:"A peer-reviewed study published in the Journal of Soil Science by researchers from a top-tier university provides detailed analysis and evidence that clay fractions in soils derived from volcanic ash predominantly contain compounds of aluminum and silicon."\\

\#\#\#\textbf{Moderately Credible Source}:"A detailed report by a well-known geologist in a respected geology magazine discusses the mineral composition of clay fractions in soils and highlights volcanic ash as a common origin of aluminum and silicon compounds."\\

\#\#\#\textbf{Low Credibility Source}:"A gardening enthusiast's blog post mentions that soils rich in aluminum and silicon compounds often come from volcanic ash, based on their personal observations and informal tests."\\

\#\#\#\textbf{question}: \{question\}\\
\#\#\#\textbf{answer}: \{answer\}\\
\\
For this \#\#\#question, \#\#\#answer pairs, make 3 evidences with difference power of evidence in the aspect of Source Credibility.

\end{tcolorbox}
\caption{The prompt for generating various of evidence according to credibility. We did not use moderate credibility evidence, as it is similar to other evidence.}
\label{tab:t5_1}
\end{table}

\begin{table}[htb]
\begin{tcolorbox}[toprule=1mm,colback=gray!10!white,colframe=gray!50,title= Prompt for Making specificity evidence, fonttitle=\bfseries]
\#\#\#\textbf{question}: "What substance does the phillosopher stone change the base material to?"\\
\#\#\#\textbf{answer}:  "gold"\\

For this \#\#\#question, \#\#\#answer pairs, make 3 evidences with difference power of evidence in the aspect of Specificity and detail.\\

\#\#\#\textbf{Highly Specific Evidence}: "Detailed records from 16th-century experiments show precise measurements and procedures for transmuting lead into gold using a substance identified as the Philosopher's Stone."\\

\#\#\#\textbf{Moderately Specific Evidence}: "Historical documents suggest that some alchemists reported converting metals into gold, but the details are sparse."\\

\#\#\#\textbf{General Evidence}: "There are general mentions in ancient texts about the ability to convert base metals into gold."\\

\#\#\#\textbf{question}: "Compounds with aluminum and silicon are commonly found in the clay fractions of soils derived from what?"\\
\#\#\#\textbf{answer}:  "volcanic ash"\\

For this \#\#\#question, \#\#\#answer pairs, make 3 evidences with difference power of evidence in the aspect of Specificity and detail.\\

\#\#\#\textbf{Highly Specific Evidence}:"Geochemical analyses of soil samples from regions with known volcanic activity demonstrate that the clay fractions are predominantly composed of alumino-silicate minerals, confirming that these soils are derived from volcanic ash deposits."\\

\#\#\#\textbf{Moderately Specific Evidence}:"Scientific studies indicate that soils in volcanic regions frequently contain clay fractions rich in aluminum and silicon compounds, which suggests a derivation from volcanic ash."\\

\#\#\#\textbf{General Evidence}:"Many references in soil science literature mention that clay fractions with aluminum and silicon are often associated with volcanic ash origins."\\

\#\#\#\textbf{question}: \{question\}\\
\#\#\#\textbf{answer}: \{answer\}\\

For this \#\#\#question, \#\#\#answer pairs, make 3 evidences with difference power of evidence in the aspect of Specificity and detail.\\

\end{tcolorbox}
\caption{The prompt for generating various evidence according to specificity. We did not use moderate specific evidence, as it is similar to other evidence}
\label{tab:t5_1}
\end{table}

\begin{table}[htb]
\begin{tcolorbox}[toprule=1mm,colback=gray!10!white,colframe=gray!50,title= Prompt for Making timeliness evidence, fonttitle=\bfseries]
\#\#\#\textbf{question}: "What substance does the phillosopher stone change the base material to?"\\
\#\#\#\textbf{answer}:  "gold"\\

For this \#\#\#question, \#\#\#answer pairs, make 2 evidences with difference power of evidence in the aspect of timeliness. (the older evidence should be before 18th-century)\\

\#\#\#\textbf{Recent Evidence}: "A 2022 study published in a scientific journal provides new experimental data supporting the possibility of metal transmutation using a newly synthesized substance resembling the Philosopher's Stone."\\

\#\#\#\textbf{Older Evidence}: "A 17th-century manuscript claims to have witnessed the transformation of base metals into gold using an alchemical process."\\

\#\#\#\textbf{question}: "Compounds with aluminum and silicon are commonly found in the clay fractions of soils derived from what?"\\
\#\#\#\textbf{answer}:  "volcanic ash"\\

For this \#\#\#question, \#\#\#answer pairs, make 2 evidences with difference power of evidence in the aspect of timeliness. (the older evidence should be before 18th-century)\\

\#\#\#\textbf{Recent Evidence}: "A 2019 study published in a geochemistry journal confirms that soils derived from volcanic ash predominantly contain clay fractions with high concentrations of aluminum and silicon compounds."\\

\#\#\#\textbf{Older Evidence}: "A 16th-century agricultural text describes soils from regions with volcanic activity as rich in aluminosilicate clays, derived from the weathering of volcanic ash."\\

\#\#\#\textbf{question}: \{question\}\\
\#\#\#\textbf{answer}: \{answer\}\\

For this \#\#\#question, \#\#\#answer pairs, make 2 evidences with difference power of evidence in the aspect of timeliness. (the older evidence should be before 18th-century)

\end{tcolorbox}
\caption{The prompt for generating various evidence according to timeliness.}
\label{tab:t5_1}
\end{table}

\begin{table}[htb]
\begin{tcolorbox}[toprule=1mm,colback=gray!10!white,colframe=gray!50,title= Prompt for Making experimental evidence, fonttitle=\bfseries]
\#\#\#\textbf{question}: "What substance does the phillosopher stone change the base material to?"
\#\#\#\textbf{answer}:  "gold"

For this \#\#\#question, \#\#\#answer pairs, make 2 evidences with different levels of strength in the aspect of Experimental or Observational Evidence, ensuring that the observational evidence includes direct observations from normal people such as "several witnesses observed."\\

\#\#\#\textbf{Experimental Evidence}: "Recent laboratory experiments conducted under controlled conditions have demonstrated the conversion of lead into gold using a synthetic version of the Philosopher's Stone."\\

\#\#\#\textbf{Observational Evidence}: "Several eyewitness accounts from the 1600s describe seeing alchemists successfully convert metals into gold, though these were not scientifically verified."\\

\#\#\#\textbf{question}: "Compounds with aluminum and silicon are commonly found in the clay fractions of soils derived from what?"\\
\#\#\#\textbf{answer}:  "volcanic ash"\\

For this \#\#\#question, \#\#\#answer pairs, make 2 evidences with different levels of strength in the aspect of Experimental or Observational Evidence, ensuring that the observational evidence includes direct observations from normal people such as "several witnesses observed."\\

\#\#\#\textbf{Experimental Evidence}: "A series of controlled soil analysis experiments have shown that soils formed from volcanic ash consistently contain high concentrations of aluminum and silicon compounds in their clay fractions."\\

\#\#\#\textbf{Observational Evidence}: "Several teams have directly observed that soils in regions with volcanic activity, particularly those rich in clay, contain significant amounts of aluminum and silicon."\\
\#\#\#\textbf{question}: \{question\}\\
\#\#\#\textbf{answer}: \{answer\}\\

For this \#\#\#question, \#\#\#answer pairs, make 2 evidences with different levels of strength in the aspect of Experimental or Observational Evidence, ensuring that the observational evidence includes direct observations from normal people such as "several witnesses observed."

\end{tcolorbox}
\caption{The prompt for generating various evidence according to the existence of the experiment. }
\label{tab:t5_1}
\end{table}

\clearpage
\bigbreak

\end{document}